\documentclass{ieeeaccess}

\usepackage[utf8]{inputenc}
\usepackage[T1]{fontenc}
\usepackage{amsmath,amssymb}
\usepackage{courier}
\usepackage{graphicx}
\usepackage{textcomp}

\usepackage{bm}
\makeatletter
\AtBeginDocument{\DeclareMathVersion{bold}
\SetSymbolFont{operators}{bold}{T1}{times}{b}{n}
\SetSymbolFont{NewLetters}{bold}{T1}{times}{b}{it}
\SetMathAlphabet{\mathrm}{bold}{T1}{times}{b}{n}
\SetMathAlphabet{\mathit}{bold}{T1}{times}{b}{it}
\SetMathAlphabet{\mathbf}{bold}{T1}{times}{b}{n}
\SetMathAlphabet{\mathtt}{bold}{OT1}{pcr}{b}{n}
\SetSymbolFont{symbols}{bold}{OMS}{cmsy}{b}{n}
\renewcommand\boldmath{\@nomath\boldmath\mathversion{bold}}}
\makeatother

\usepackage{microtype}
\usepackage{url}
\usepackage[hidelinks]{hyperref}
\usepackage{enumitem}
\usepackage[caption=false,font=scriptsize]{subfig}
\usepackage{booktabs}
\usepackage{makecell}
\usepackage{algorithm}
\usepackage[noend]{algpseudocode}
\usepackage{float}

\usepackage[abbreviate=true, dateabbrev=true, alldates=year, isbn=true, doi=true, urldate=comp, url=false, maxbibnames=9, backref=false, backend=biber, style=numeric-comp, language=american, giveninits=true, sortcites=true, sorting=none, natbib=false]{biblatex}
\usepackage{citation-cleaner}
\AtEveryBibitem{\clearlist{location}} %
\AtEveryBibitem{\clearname{editor}} %
\AtEveryBibitem{\clearfield{issn}} %
\AtEveryBibitem{\clearfield{series}} %
\AtEveryBibitem{\clearlist{language}} %
\AtEveryBibitem{\clearfield{note}} %
\RemoveArxivDOIs{} %
\AtEveryBibitem{\iffieldundef{doi}{}{\clearfield{isbn}}} %

\usepackage{framed} 
\definecolor{shadecolor}{gray}{0.9}
\newenvironment{rqanswer}{%
  \setlength{\OuterFrameSep}{3pt}%
  \MakeFramed {\advance\hsize-\width \FrameRestore}}%
 {\endMakeFramed}

\newcommand{\takeaway}[2]{%
  \begin{rqanswer}%
    \head{#1:}\ #2%
  \end{rqanswer}%
}

\newcommand{\head}[1]{\par\noindent\textbf{#1}\space}
\newcommand{\floatfont}{\footnotesize}

\algrenewcommand\algorithmicrequire{\textbf{Input:}}
\algrenewcommand\algorithmicensure{\textbf{Output:}}
\algrenewcommand\algorithmiccomment[1]{\hfill$\triangleright$~#1}

\def\headerlogo{\rule[-\baselineskip]{0pt}{3\baselineskip}}
\def\headerlogoall{\headerlogo}%

\usepackage{etoolbox}%

\newlength\extramargin
\begin{document}

\history{}
\def\theyear{2026}
\def\thevol{0}
\def\doifont#1\vss{} %
\makeatletter %
\patchcmd{\ps@titlepage}%
  {\footervolfont VOLUME\ \thevol, \theyear}{\mycopyrightnotice}{}{}
\patchcmd{\ps@titlepage}%
  {\footervolfont VOLUME\ \thevol, \theyear}{\mycopyrightnotice}{}{}
\patchcmd{\ps@headings}%
  {\footervolfont VOLUME\ \thevol, \theyear}{}{}{}
\patchcmd{\ps@headings}%
  {\footervolfont VOLUME\ \thevol, \theyear}{}{}{}
\ps@headings 
\makeatother

\def\mycopyrightnotice{%
  \raisebox{-3mm}{%
    \hspace*{2.5mm}\includegraphics[width=2cm]{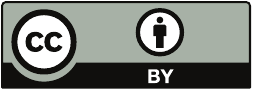}%
    \hspace*{2mm}\raisebox{2.5mm}{\footnotesize%
      \parbox{\columnwidth}{\footnotesize This work is licensed under a Creative Commons \\ Attribution 4.0 International (CC BY 4.0) license.}\\
      \hspace*{-17mm}\raisebox{1.6mm}{\footnotesize%
      Copyright \textcopyright\ 2026 held by the author(s).}%
    }%
  }%
  \gdef\mycopyrightnotice{}%
}

\title{Self-Healing Visual Recovery\\for Autonomous Ground Vehicles\\Using Camera-Only Visual Odometry}

 \author{%
    \uppercase{Jakob Solberg Berntzen}\authorrefmark{1,2},
    \uppercase{Safia Fatima}\authorrefmark{1,2}, AND 
    \uppercase{Leon Moonen}\authorrefmark{1} \IEEEmembership{Member, IEEE}
}
\address[1]{Simula Research Laboratory, Oslo, Norway}
\address[2]{ University of Oslo, Oslo, Norway}
\tfootnote{This work was supported in part by the Research Council of Norway through the cureIT project (grant \#300461).}

\corresp{Corresponding author: Leon Moonen (e-mail: leon@simula.no).}

\markboth
{J. S. Berntzen, S. Fatima, and L. Moonen: Self-Healing Visual Recovery for Autonomous Ground Vehicles \ldots}
{J. S. Berntzen, S. Fatima, and L. Moonen: Self-Healing Visual Recovery for Autonomous Ground Vehicles \ldots}

\begin{abstract}
Low-cost unmanned ground vehicles are often used in indoor places like warehouses, inspection corridors, and farm rows, where painted floor lines guide the robot. Line following is useful because it only needs one camera and little computing power, but it can fail when the line is blocked or turns sharply and goes out of view. Sensor-rich platforms tolerate this through hardware redundancy (LiDAR, GPS, multiple cameras), but camera-only systems must recover at runtime with no additional infrastructure. This paper presents a lightweight, two-stage recovery approach that restores guideline tracking without LiDAR, GPS, or a GPU. When the line is lost, the robot first turns in place while slowly relaxing its color checks and waiting for confirmation across multiple frames~(Stage~1). If the line is still not found, monocular visual odometry moves the robot back to saved breadcrumb positions before it tries again~(Stage~2). The system uses a depth-gated HSV line tracker, a YOLOv8n obstacle detector, and a visual odometry breadcrumb mapper, and it runs at 20 Hz on CPU-only hardware. The controller embeds a complete MAPE-K loop within a single 50\,ms control tick, with no external adaptation manager required. The approach is evaluated across 119 fault-injected episodes on three Webots simulation courses. The method was successful in 86.6\% of cases, with a median recovery time of 3.26 seconds. These results demonstrate that reliable visual recovery is feasible on camera-only UGVs within practical cost and computational limits.
\end{abstract}

\begin{keywords}
Autonomous ground vehicle, 
self-adaptive systems, 
line following, 
visual odometry, 
obstacle avoidance, 
fault tolerance, 
low-cost robotics, 
YOLOv8, 
depth-gated perception, 
Webots simulation.
\end{keywords}

\titlepgskip=-22pt

\maketitle

\section{Introduction}
\label{sec:intro}

\PARstart{A}{utonomous} ground vehicles~(UGVs) are used in a growing range of applications, including logistics, industrial inspection, and small-scale agriculture, where continuous and unattended operation is valuable~\cite{wei2024:research,agelli2024:unmanned}. Falling component costs and persistent labor shortages have moved these vehicles from laboratory prototypes into routine field use for tasks such as crop monitoring, transport, and inspection, where reliable operation must be maintained under real-world conditions~\cite{wei2024:research,agelli2024:unmanned}. Section~\ref{sec:related} examines this demand and the case for low-cost autonomy in more detail.

High-end UGVs deployed in demanding sectors such as space exploration, military reconnaissance, and precision agriculture rely on redundant sensor suites that may include LiDAR, RADAR, GPS, multiple cameras, and inertial measurement units, together with sophisticated sensor-fusion architectures for localization and fault detection~\cite{maimone2007:two,cadena2016:present,agelli2024:unmanned}. These configurations deliver reliable navigation in dynamic and unstructured environments, but their substantial acquisition cost, integration complexity, and maintenance requirements place them beyond the financial reach of many small-to-medium enterprises and developing-market operators~\cite{misaros2023:autonomous,spagnuolo2025:agricultural}.

Many sectors, including warehouse logistics, floor-guided inspection, and small-scale agriculture, depend on repetitive, labor-intensive tasks that could benefit greatly from automation but cannot justify the cost of sensor-rich platforms~\cite{misaros2023:autonomous}. Guideline following is a navigation paradigm particularly well suited to these deployments, a painted or taped floor line provides a deterministic reference with minimal infrastructure cost, and a robot can track it using only a single camera and limited onboard processing~\cite{bonin-font2008:visual}. Line following is therefore widely adopted in warehouse automation, conveyor-guided inspection robots, and agricultural row-following systems, where structured layouts make such guidance practical and cost-effective.

Camera-first UGVs provide a cost-effective alternative by relying on a single RGB-D sensor alongside modern computer vision and deep learning techniques. Recent advances in lightweight object detection~\cite{wang2023:yolov7,jocher2023:ultralytics}, monocular depth estimation~\cite{godard2019:digging,ranftl2022:robust}, and visual odometry (VO)~\cite{scaramuzza2011:visual,engel2018:direct} have made camera-only platforms increasingly practical in structured indoor environments. Even so, an important resilience gap remains: how can a camera-only system \emph{detect} navigational failure, specifically the complete loss of its primary visual reference, and \emph{recover autonomously} without external positioning aids?

The failure modes of line-following systems arise from three principal sources: (i) partial or complete occlusion of the guide line by obstacles or debris, (ii) strong reflections, lighting changes, or worn markings that degrade color segmentation reliability, and (iii) heading deviations induced by obstacle avoidance maneuvers that carry the line beyond the camera field of view. Each failure mode demands a different recovery response, a brief heading correction may suffice for a partial occlusion, while complete line loss requires a systematic search or autonomous navigation back to a previously observed pose~\cite{lee2012:visionbased,luo2013:resume}. Without an effective recovery mechanism, even a single unrecovered failure halts the robot until a human intervenes, undermining the economic justification for autonomous deployment.

Self-healing systems research~\cite{white2004:architectural,cheng2015:using} characterizes this failure mode as a runtime recovery challenge, requiring the system to detect deviations from expected behavior and apply corrective actions autonomously without human intervention. The MAPE-K (Monitor-Analyze-Plan-Execute-Knowledge) architectural pattern~\cite{white2004:architectural,delemos2013:software} has been adapted for robot navigation and service continuity~\cite{romero-garces2022:managing,misaros2023:autonomous}, but CPU-only, single-camera platforms cannot easily support the external adaptation managers or runtime environment models typically assumed by full MAPE-K implementations.

Existing recovery methods rely either on costly sensors for global localization~\cite{thrun2001:robust,cadena2016:present} or on deep learning pipelines that require dedicated GPU compute~\cite{yu2020:deeplearningbased,teed2021:droidslam}. By contrast, lightweight alternatives that run on a single CPU remain underexplored. This is especially true for guideline following, a control paradigm widely used in warehouse automation, floor-guided inspection robots, and agricultural row-following platforms where low hardware cost is paramount.

The main contributions of this paper are as follows:
\begin{enumerate}
\item We present a unified perception-control-recovery loop that operates at 20 Hz on CPU-only hardware, without relying on GPU acceleration.
\item We develop a depth-gated HSV line tracker that integrates online hue adaptation via exponential moving average (EMA) with per-row floor modeling and periodic SVD-based plane fitting to reduce false positives on reflective or cluttered surfaces.
\item We propose a depth-fused obstacle avoidance module that combines geometric residuals from a learned floor model with YOLOv8n bounding-box detections, prioritized by median depth within a forward image corridor.
\item We introduce a two-stage visual recovery strategy: first, an in-place spin-and-search procedure with relaxed thresholds and strict multi-frame confirmation; second, VO-guided navigation to stored breadcrumb poses when the initial search fails.
\item We conduct an empirical evaluation across 119 induced line-loss episodes on three geometrically diverse courses.
\end{enumerate}

The rest of the paper is organized as follows. Foundational concepts are introduced in Section~\ref{sec:background}. Related work is reviewed in Section~\ref{sec:related}.
The system design is presented in Section~\ref{sec:method}, and the experimental
setup is described in Section~\ref{sec:eval}. Results and discussion are presented in
Section~\ref{sec:results}, threats to validity in Section~\ref{sec:threats},
future directions in Section~\ref{sec:future}, and the conclusion in Section~\ref{sec:conclusion}.

\section{Background}
\label{sec:background}

\subsection{Autonomous Ground Vehicle Navigation}

Autonomous robots execute a continuous sense--decide--act cycle: sensors capture the environment, an onboard decision process selects an appropriate action, and actuators carry it out~\cite{thrun2002:probabilistic}. Because sensors are imperfect and actuation is noisy, modern robotic systems adopt a probabilistic view that models both state and observation uncertainty explicitly. Bayesian filtering provides the core mechanism, a prior belief over the vehicle state is updated at each timestep by combining a motion model $p(x_k \mid x_{k-1}, u_k)$ with a sensor observation model $p(z_k \mid x_k)$, yielding a posterior that degrades gracefully rather than failing catastrophically when sensor data is sparse or noisy. This probabilistic formulation is especially valuable for low-cost platforms, where individual sensors may be unreliable.

For line-following UGVs, the relevant state is the lateral offset and heading relative to the guide line. When this visual reference is available, a proportional-derivative (PD) controller is sufficient for stable tracking. When the reference signal is lost (for example, in the absence of a fallback mechanism), the controller no longer receives a valid input. As a result, the vehicle may come to a stop or drift indefinitely. Self-healing behavior therefore requires the system to detect loss autonomously, execute a recovery strategy, and reacquire the reference without human intervention.

\subsection{Visual Odometry}

Visual odometry (VO) estimates incremental camera motion from a sequence of images without GPS or wheel encoders~\cite{scaramuzza2011:visual,nister2004:visual}. A typical monocular VO pipeline consists of three stages: (i)feature detection, using detectors such as GFTT or ORB to identify salient keypoints, (ii)feature matching across frames, with KNN and Lowe's ratio test to filter ambiguous correspondences, and (iii) motion recovery, using a five-point algorithm inside RANSAC to solve for the essential matrix $\mathbf{E}$, which encodes the relative rotation $\mathbf{R}$ and translation direction $\hat{\mathbf{t}}$ between frames.

A key limitation is drift, as each pose estimate is conditioned on the previous one, small errors accumulate over time and push the trajectory away from the true path, especially in low-texture or repetitive environments. Full SLAM systems correct drift via loop closure and bundle adjustment~\cite{cadena2016:present}, but these operations are too costly for CPU-only, real-time control. For short-range navigation (under 5\,m), however, a lightweight GFTT along with ORB and five-point pipeline provides sufficient pose accuracy without a heavy SLAM back-end, making it well suited to the breadcrumb-based recovery strategy described in this paper.

\subsection{MAPE-K Self-Adaptive Architecture}

The MAPE-K architectural pattern~\cite{white2004:architectural,cheng2015:using} follows runtime adaptation using four sequential stages operating over a shared knowledge base: \emph{Monitor} collects system and environment observations, \emph{Analyze} detects deviations from expected behavior, \emph{Plan} selects a corrective strategy, and \emph{Execute} applies the chosen action through effectors. In canonical deployments, these stages are implemented as a separate adaptation manager that communicates with managed components through well-defined interfaces, allowing the adaptation logic to be independently developed and verified~\cite{delemos2013:software}.

For embedded robotics on CPU-only hardware, a separate adaptation manager adds too much latency and resource overhead. The controller in this paper instead runs the MAPE-K loop in line with the control cycle, rather than through an external manager. This gives up the modularity of a full MAPE-K deployment, but keeps latency and resource use low enough for the low-cost UGV hardware targeted here. Section~\ref{sec:method} describes how the four control phases carry out the MAPE-K loop.

\section{Related Work} \label{sec:related}

\subsection{Industrial Demand and the Case for Affordable Autonomy}

The growth of service and industrial robotics creates strong economic pressure to reduce the cost of autonomous systems. The IFR World Robotics 2025 report puts the worldwide operational stock of industrial robots at 4.66 million units in 2024, a 9\% rise over the previous year, with 542,000 new units installed~\cite{robotics:sales2025}. In the same year, nearly 200,000 professional service robots were sold, again up 9\%, of which transportation and logistics accounted for the largest share at 102,900 units (up 14\%)~\cite{robotics:service2025}. Robot density remains high in leading markets, reaching 1,220 units per 10,000 employees in the Republic of Korea, 818 in Singapore, and 449 in Germany~\cite{robotics:global2025}. This expansion is driven not only by large manufacturers but increasingly by small and medium enterprises seeking to automate repetitive tasks at lower capital investment~\cite{misaros2023:autonomous,spagnuolo2025:agricultural}.

High-end UGVs achieve robustness through sensor redundancy and hardware diversity, but their cost and complexity limit adoption~\cite{liu2020:survey,agelli2024:unmanned}. Camera-based platforms address this gap by replacing expensive sensors with computer vision and machine learning~\cite{mittal2024:comprehensive}. The challenge that remains is reliability, unlike sensor-rich systems that tolerate individual sensor failures through redundancy, camera-first systems must handle failure modes through algorithmic resilience rather than hardware redundancy. This motivates the design of embedded recovery strategies that can operate entirely on commodity CPU hardware.

\subsection{Visual Odometry and SLAM}

Visual odometry (VO) estimates camera motion by tracking feature correspondences across successive frames~\cite{scaramuzza2011:visual,nister2004:visual}. The field can be broadly divided into two families of approaches based on how they extract geometric information from images.

\begin{figure}[t]
  \centering
  \includegraphics[width=\linewidth]{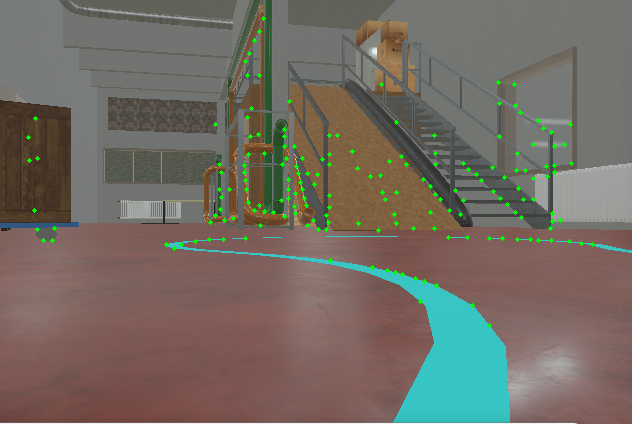}
  \vspace*{-2ex}
  \caption{Feature based Visual SLAM: tracked keypoints (green) overlaid on the image and the estimated camera trajectory (blue) accumulated from matched feature correspondences across keyframes. Drift accumulates along the trajectory unless corrected by loop closure and bundle adjustment.}
  \label{fig:vslam}
\end{figure}

\emph{Feature-based visual SLAM} has been a central research direction in robot perception because it offers an interpretable pipeline for estimating motion and building a sparse map from images. These methods detect salient keypoints (for example, FAST corners) and compute compact descriptors (e.g., ORB) that can be matched across frames using distance metrics and ratio tests to infer inter-frame correspondences. A landmark system in this line of work is PTAM~\cite{klein2007:parallel}, which demonstrated real-time monocular SLAM by splitting tracking and mapping into parallel threads, enabling a stable front-end pose estimate while the back-end incrementally optimizes a keyframe map. Building on this architecture, the ORB-SLAM family~\cite{mur-artal2015:orbslam,campos2021:orbslam3} introduced more robust relocalization, loop-closure detection using bag-of-words techniques~\cite{galvez-lopez2012:bags}, and support for monocular, stereo, and RGB-D sensing. ORB-SLAM3 further extended the framework to visual--inertial configurations and a multi-map design, establishing a widely used baseline for feature-based V-SLAM~\cite{campos2021:orbslam3}. Since relative pose estimates inevitably accumulate error over long trajectories, state-of-the-art feature-based systems reduce drift through global bundle adjustment and pose-graph optimization in factor-graph back-ends~\cite{cadena2016:present,kaess2012:isam2}. Figure~\ref{fig:vslam} summarizes the typical outputs of such a pipeline, including tracked keypoints and the resulting estimated camera trajectory.

\emph{Direct methods} avoid explicit keypoint extraction and instead minimize photometric alignment error over high-gradient pixels, making them accurate over short baselines when the brightness constancy assumption holds~\cite{engel2014:lsdslam,engel2018:direct}. LSD-SLAM~\cite{engel2014:lsdslam} extended this approach to large scale, semi-dense mapping, while Direct Sparse Odometry (DSO)~\cite{engel2018:direct} added a fully photometric calibration model and joint optimization of poses and inverse depths. Direct methods can be very precise at short ranges but are more sensitive to lighting changes and motion blur than feature based alternatives.

\emph{Visual-inertial odometry (VIO)} tightly couples image measurements with IMU pre-integration to stabilize motion estimates during rapid turns, aggressive maneuvers, or weak texture regions, and to recover metric scale from monocular input~\cite{forster2017:onmanifold}. OKVIS~\cite{leutenegger2015:keyframebased} introduced the tightly coupled non-linear optimization formulation, while VINS-Mono~\cite{qin2018:vinsmono} demonstrated robust monocular VIO on embedded platforms using a sliding-window smoother and loop closure. These systems substantially improve robustness at the cost of requiring an IMU, which adds hardware and calibration complexity.

\emph{Learned front-ends} replace hand crafted feature descriptors with differentiable correspondence networks. DROID-SLAM~\cite{teed2021:droidslam} formulates tracking and mapping as iterative updates to a dense flow field using recurrent neural networks, while DPVO~\cite{teed2023:deep} extends this to sparse patch correspondences for higher throughput. Surveys of deep V-SLAM report a growing trend to combine learned place recognition with geometric back-ends for better performance under strong appearance change~\cite{favorskaya2023:deep}, but the GPU requirement of these networks precludes their use on low-cost CPU-only platforms.

Most V-SLAM systems separate the estimation problem into a front end and a back end~\cite{cadena2016:present}. The front end extracts geometric constraints, the back end optimizes a factor graph using bundle adjustment and pose-graph optimization, with incremental solvers such as iSAM2~\cite{kaess2012:isam2} exploiting sparsity to update solutions efficiently as new observations arrive. In practice, the choice of front-end family involves a trade-off: feature based methods handle large viewpoint changes and support strong loop closure while direct methods are precise at short baselines but sensitive to lighting. VIO adds short-term stability at the cost of additional hardware and learned front-ends improve robustness at the cost of GPU inference.

Two-stage recovery in this work requires only short-range VO (typically under 5\,m of travel to a stored pose). A GFTT along with ORB and five-point pipeline provides sufficient accuracy for this purpose without the overhead of a full SLAM back-end or a GPU-dependent learned front-end.

\subsection{Displacement Recovery and Fault-Tolerant Navigation}

Displacement recovery refers to cases in which a robot is physically moved while its internal state estimate remains unchanged, causing the belief to become confident but inconsistent with actual sensor observations~\cite{thrun2002:probabilistic,thrun2001:robust}. The system must quickly detect this inconsistency and restore a correct global pose before normal operation can resume.

\emph{Probabilistic approaches} address kidnapping through adaptive particle filters~\cite{fox2001:kldsampling,thrun2001:robust} that expand or contract the particle population based on the correspondence between predicted and observed measurements. When sensor-state consistency drops below a threshold, the filter injects additional hypotheses to cover a wider region of the state space, then prunes back once a consistent pose is found. This approach gracefully handles sudden localization failures without requiring a global map query, but its scalability is limited by the number of particles needed to cover high-dimensional state spaces.

Appearance-based place recognition is widely used in visual SLAM to generate candidate poses for recovery after tracking failure. In bag-of-words methods such as Galvez’s approach~\cite{galvez-lopez2012:bags}, local image descriptors are quantized into a learned visual vocabulary, and keyframes are represented by histograms over these visual words. This representation enables fast retrieval of keyframes with similar appearance from large databases, providing effective relocalization hypotheses for subsequent geometric verification. FAB-MAP~\cite{cummins2008:fabmap} extends this idea with a probabilistic generative model of visual scene appearance, which explicitly accounts for the probability of observing visual words given a particular location and improves robustness under lighting and viewpoint change. Retrieved candidates are geometrically verified using PnP estimation or a five-point essential matrix solver inside RANSAC before being accepted, and successful relocalization poses are fused into the SLAM back-end through pose-graph optimization~\cite{mur-artal2015:orbslam}.

\emph{Geometry-based descriptors} provide an alternative when visual appearance is ambiguous or repeat-textured. ScanContext~\cite{kim2018:scan} encodes the spatial distribution of a LiDAR point cloud into a compact 2D array indexed by range and angle, enabling efficient place retrieval without the sensitivity to lighting that affects appearance-based methods. SegMatch~\cite{dube2017:segmatch} further segments the 3D scene into objects and matches segments across different traversals, reducing the impact of dynamic changes. These descriptors reduce the candidate search space through geometric consistency checks before expensive alignment, speeding up recovery and lowering false-positive rates in repetitive environments.

\emph{Learning-based relocalization} offers a complementary strategy. Regression models such as PoseNet~\cite{kendall2015:posenet} directly predict a 6-DOF camera pose from a single image using a convolutional network trained on a specific scene, providing a coarse absolute-pose estimate in constant time without an explicit map query. Retrieval models such as NetVLAD~\cite{arandjelovic2016:netvlad} encode image content into a compact global descriptor optimized for place retrieval, producing strong candidates that are then geometrically refined. Both approaches require GPU inference and scene-specific training, limiting their applicability to the camera-only, CPU-only target of this work.

Prior work has also demonstrated embedded visual recovery on resource-constrained hardware without these heavier components. Work in~\cite{lee2012:visionbased} reported 87\% qualitative success and a Mean Time to Recovery (MTTR) of approximately 3.23\,s on an ARM11 processor. Another system combined appearance cues with motion primitives, reducing MTTR by roughly half relative to Monte Carlo Localization~\cite{luo2013:resume}. More recently, deep learning has been applied to failure detection in autonomous driving, improving robustness under distribution shift while requiring GPU inference~\cite{yu2020:deeplearningbased}. These results confirm that competitive recovery performance is achievable at lower hardware tiers, motivating the approach presented in this paper.

These recovery stages form a MAPE-K loop~\cite{white2004:architectural,cheng2015:using,delemos2013:software}; Section~\ref{sec:method} describes how the control phases realize its stages. Probabilistic model checkers such as PRISM~\cite{calinescu2011:dynamic,kwiatkowska2011:prism} can verify recovery guarantees under abstracted environment models, but constructing such models is too costly for real-time control loops; this work therefore relies on empirical MTTR and success-rate metrics.

\subsection{Lightweight AI for Robotic Perception}

Deep learning has substantially broadened the capabilities of vision-based robotic systems by providing both semantic understanding and geometric cues at low latency.

\emph{Object detection} networks such as YOLOv7 and YOLOv8 formulate detection as a single grid-based regression problem, dividing the input image into a spatial grid and directly predicting bounding boxes and class probabilities for each cell~\cite{wang2023:yolov7,jocher2023:ultralytics}. This unified formulation avoids the two-stage region-proposal overhead of earlier detectors and achieves real-time throughput competitive with slower, more accurate architectures. These models are typically trained on large-scale datasets such as COCO~\cite{lin2014:microsoft}, which provides 80 object categories and over 330,000 images, giving the network the visual diversity needed for robust generalization. Exporting Ultralytics models to ONNX format and deploying them via the OpenCV DNN backend enables CPU-only inference on ARM and x86 platforms without requiring a dedicated GPU or proprietary acceleration library~\cite{mallik2022:realtime}. A practical demonstration of this approach is a monocular obstacle avoidance pipeline that exports a RetinaNet-50 model to ONNX and compiles it with TensorRT on Jetson-class hardware, confirming that deep detection models can be adapted for embedded platforms under constrained resources~\cite{mallik2022:realtime}.

\emph{Semantic segmentation} networks parse the full image at the pixel level and are useful for identifying floor regions, free space, and obstacles simultaneously. Lightweight architectures designed for embedded deployment include ENet~\cite{paszke2016:enet}, which reduces computation through early downsampling and asymmetric convolutions; BiSeNet~\cite{yu2018:bisenet}, which maintains spatial resolution through a bilateral branch structure; and Fast-SCNN~\cite{poudel2019:fastscnn}, which achieves real-time throughput through a shared feature extraction layer and a lightweight decoder. These networks enable robots to interpret scene layout continuously without overwhelming limited onboard processing budgets, making them suitable for floor detection and free-space mapping on mobile platforms.

\emph{Monocular depth estimation} provides complementary range information when a dedicated depth sensor is unavailable or when its accuracy is insufficient for fine-grained tasks. Monodepth2~\cite{godard2019:digging} uses self-supervised training from stereo pairs or monocular sequences with a novel masking strategy to handle moving objects, while MiDaS~\cite{ranftl2022:robust} trains on a diverse mixture of depth datasets to produce relative-depth estimates that generalize across scene types. These methods are primarily useful for tasks that require qualitative depth ordering rather than metric accuracy.

In this paper, the system uses a depth sensor directly for metric range measurements, so monocular depth estimation is not needed. YOLOv8n (the nano variant) is used as a semantic complement to a geometric depth-residual obstacle detector. The two detection modalities are fused during the Analyze phase, and candidate obstacles are ranked by their median depth within the forward image corridor, giving priority to the closest detected object regardless of whether it was identified by the geometric or the semantic channel.

\subsection{Self-Adaptive Systems}

Self-adaptive systems are designed to modify their own behavior at runtime in response to changes in the environment or in the system itself, without human intervention~\cite{white2004:architectural,cheng2015:using}. The MAPE-K architectural pattern~\cite{white2004:architectural,delemos2013:software} is the most widely adopted framework for organizing this adaptation: \emph{Monitor} observes system state and environmental conditions; \emph{Analyze} detects deviations from expected behavior using the monitored data; \emph{Plan} selects and sequences corrective strategies; and \emph{Execute} applies the chosen plan through actuators or configuration changes, all over a shared \emph{Knowledge Base} that persists runtime state across loop iterations.

The Rainbow framework~\cite{cheng2015:using} was one of the first platforms to realize MAPE-K in software systems, introducing the concept of an external adaptation manager that governs managed application components through a model-based interface. Subsequent work in the robotics domain has adapted this structure to navigation and service-continuity scenarios, demonstrating that self-adaptive loops can improve predictability, enhance safety, and strengthen operational resilience during runtime faults~\cite{romero-garces2022:managing}. A key challenge in applying MAPE-K to robotic platforms is latency: conventional MAPE-K deployments assume asynchronous, loosely-coupled adaptation that is unsuitable for tightly-timed control loops running at 20--50\,Hz.

Recovery effectiveness in self-adaptive systems is typically quantified using time-to-recovery and recovery-success-rate metrics~\cite{beyer2016:site,romero-garces2022:managing}. 
These capture both responsiveness (how quickly recovery occurs) and reliability (how consistently it succeeds). 
We use the same metrics to evaluate our approach, and discuss them in more detail in Section~\ref{sec:metrics}.

Probabilistic model checkers such as PRISM~\cite{calinescu2011:dynamic,kwiatkowska2011:prism} can verify recovery guarantees by abstracting the adaptive system as a Markov decision process and checking temporal logic properties over all reachable states. 
Because the recovery algorithm in this paper uses explicit stage budgets, timeouts, and confirmation thresholds, it is structurally well suited to such formal analysis. 

\subsection{Research Gap}

Existing recovery methods typically rely either on redundant sensing modalities, such as LiDAR, GPS, and RADAR~\cite{thrun2001:robust,cadena2016:present}, or on GPU-class hardware for real-time inference~\cite{yu2020:deeplearningbased,campos2021:orbslam3}. Neither requirement is well suited to the growing class of low-cost, camera-first UGVs used in logistics and inspection. Although prior camera-only systems~\cite{lee2012:visionbased} have achieved comparable recovery performance, they do not incorporate onboard depth-obstacle fusion or VO-based breadcrumb navigation, which makes them more vulnerable to complete line loss in geometrically complex environments.
To the best of our knowledge, no prior work addresses all three constraints simultaneously: (i) camera-only sensing without LiDAR, RADAR, or GPS; (ii) CPU-only computation without GPU acceleration; and (iii) autonomous line-loss recovery using VO-guided return to stored poses. The proposed algorithm operates under all three constraints while achieving recovery accuracy comparable to that of heavier prior systems.

Table~\ref{tab:positioning} maps these gaps against representative related work, spanning a lightweight camera-only recovery system~\cite{lee2012:visionbased}, a deep-learning failure detector~\cite{yu2020:deeplearningbased}, and a full visual SLAM pipeline~\cite{campos2021:orbslam3}. This is a qualitative comparison of capabilities; the SLAM and deep-learning systems are included because they represent the heavier alternatives this work aims to avoid, even though they do not report line-loss recovery times.

\begin{table}[t]

\centering\floatfont
\caption{Capability analysis: proposed approach vs.\ relevant related
work consisting of a lightweight camera-only recovery system~\cite{lee2012:visionbased}, 
a deep- learning-based failure detector~\cite{yu2020:deeplearningbased}, and a full visual SLAM pipeline~\cite{campos2021:orbslam3}.
Legend: \checkmark = feature present, $\circ$ = partial, $\times$ = absent.}
\label{tab:positioning}
\vspace*{-1ex}
\begin{tabular}{lcccc}
\toprule
 & \thead{Proposed} & \thead{\cite{lee2012:visionbased}} & \thead{\cite{yu2020:deeplearningbased}} & \thead{\cite{campos2021:orbslam3}} \\
\midrule
CPU-only              & \checkmark & \checkmark & $\times$ & $\circ$ \\
No LiDAR/GPS          & \checkmark & $\circ$    & $\times$ & \checkmark \\
Line-loss recovery    & \checkmark & \checkmark & \checkmark & $\times$ \\
VO breadcrumbs        & \checkmark & $\times$   & $\times$ & $\times$ \\
Depth obstacle fusion & \checkmark & $\times$   & $\times$ & $\times$ \\
$>$85\% recovery rate & \checkmark & \checkmark & \checkmark & N/A \\
\bottomrule
\end{tabular}
\end{table}

\section{Proposed Approach}
\label{sec:method}

\begin{figure*}[t]
  \centering
  \includegraphics[width=\textwidth]{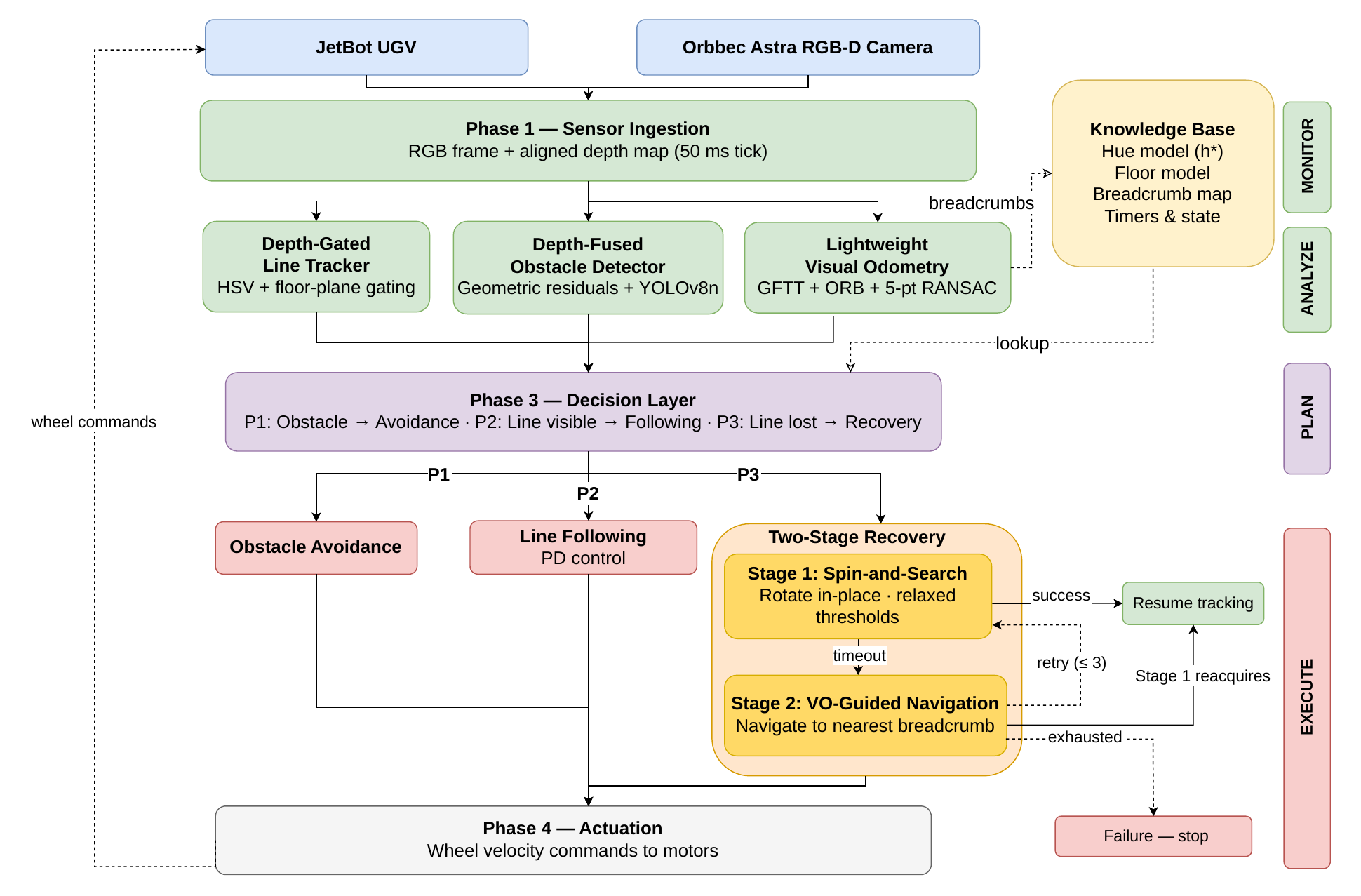}
  \vspace*{-2ex}
  \caption{Overview of the proposed recovery framework. Each 50\,ms tick traverses four phases (Monitor $\to$ Analyze $\to$ Plan $\to$ Execute) organized as a MAPE-K loop. Three perception modules feed a shared Knowledge Base: the Decision Layer selects one behavior per tick by priority (P1:~Avoidance $\succ$ P2:~Line~Following $\succ$ P3:~Recovery).}
  \label{fig:framework}
\end{figure*}

The overview of the proposed system is illustrated in Figure~\ref{fig:framework}. Each
50\,ms control tick proceeds through four sequential phases that map directly
onto a MAPE-K self-adaptive loop~\cite{white2004:architectural}: \emph{Sensor
Ingestion}~(Monitor), \emph{State Estimation}~(Analyze), \emph{Decision
Layer}~(Plan), and \emph{Actuation}~(Execute). A shared \emph{Knowledge
Base} stores the hue model~$h^*$, the ground-plane estimate, the breadcrumb
map, and all runtime timers across ticks. The following subsections describe
each phase in detail.

\subsection{Hardware Platform} \label{subsec:hardware}

\begin{figure}[t]
  \centering\tableheadfont
  \subfloat[\raggedright\tableheadfont Robot platform: the JetBot differential-drive UGV (top view).]{%
    \includegraphics[width=0.46\linewidth]{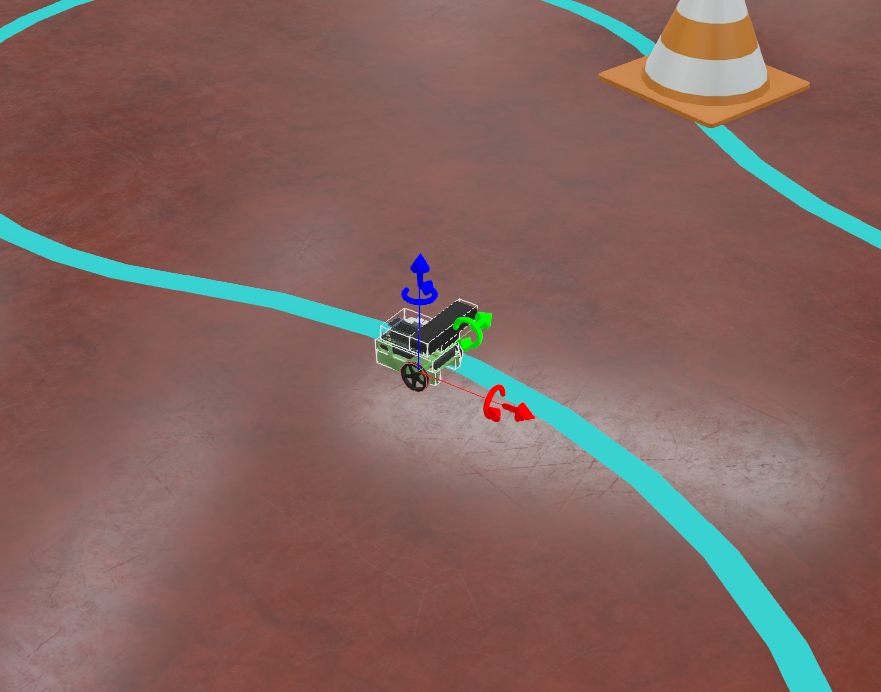}}
  \hspace{6pt}%
  \subfloat[\tableheadfont Orbbec Astra RGB-D camera providing color and depth at 30\,Hz.]{%
    \includegraphics[width=0.46\linewidth]{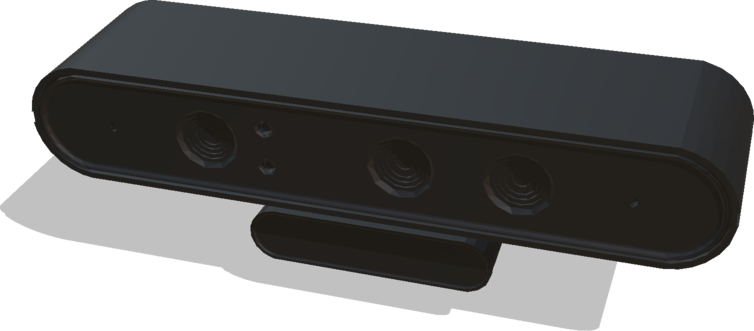}}
  \caption{Hardware configuration. The Astra camera is front-facing,
    providing synchronized RGB and depth at $640\times480$, $\approx58^\circ$ HFoV.}
  \label{fig:hardware}
\end{figure}

The target hardware is a JetBot-class differential-drive UGV equipped with an Orbbec Astra RGB-D camera (top row of Figure~\ref{fig:framework}; a closer view is shown in Figure~\ref{fig:hardware}). The Astra provides synchronized color and depth images at up to $640\times480$ resolution and 30\,Hz, with a horizontal field of view of approximately $58^{\circ}$. The depth stream gives per-pixel range measurements without a separate laser or ultrasonic sensor.

The controller is implemented in Python and combines OpenCV for image
processing, visual odometry, object detection, and depth handling; NumPy for
array operations, linear algebra, and robust statistics; and the Webots Python
API for hardware abstraction. A single YOLOv8n ONNX model is loaded through
\verb|cv2.dnn.readNetFromONNX| and run entirely on the CPU backend. The system
does not rely on a GPU, ROS, or any external middleware.

\subsection{Phase 1: Sensor Ingestion (Monitor)}
\label{subsec:sensor}

At the beginning of each control tick (Phase~1, Figure~\ref{fig:framework}), a
synchronized RGB frame and depth map are acquired from the Webots API. The
depth image is resized to $640\times480$ using nearest-neighbour interpolation
to preserve metric consistency across depth discontinuities. These aligned
images serve as the only inputs to the downstream pipeline; no GPS, LiDAR, or
wheel-encoder data are used.

\subsection{Phase 2: State Estimation (Analyze)} \label{subsec:state_est}

In Phase~2, three perception modules are executed for each new RGB-D frame pair (Figure~\ref{fig:framework}). Their outputs are combined in the shared Knowledge Base, which stores the online hue model~$h^*$, the per-row floor depth model ~$\{g_r\}$ , the breadcrumb map \verb|line_map|, and the runtime timers used by subsequent control stages.

\subsubsection{Depth-Gated Line Tracker} \label{subsec:line_tracker}

The \emph{Depth-Gated Line Tracker} (Figure~\ref{fig:framework}) detects the guide line in HSV space by combining color segmentation with a depth-aware constraint. A hue reference $h^*$ is initialized from a small seed region near the lower center of the frame and is then updated online using an exponential moving average:
\begin{equation}
  h^*_k = (1-\alpha)\,h^*_{k-1} + \alpha\,\hat{h}_k,
  \label{eq:ema}
\end{equation}
where $\hat{h}_k$ denotes the median hue within the seed region and $\alpha$ controls the adaptation rate. In normal operation, the near-field mask is generated using \verb|inRange|($h^*-\delta_h$ and $h^*+\delta_h$), where $\delta_h$ sets the allowable hue tolerance.

Raw color segmentation alone is prone to false positives on reflective floor surfaces and non-floor objects. To improve robustness, we apply a depth-based floor mask. Each image row $r$ maintains an exponential moving average (EMA) of observed depth values:
\begin{equation}
g_r^{(k)} = (1-\alpha_g)\,g_r^{(k-1)} + \alpha_g\,D_r^{(k)},
\label{eq:row_ema}
\end{equation}
with $\alpha_g = 0.05$ and a 20-frame initialization warmup. Every three frames, we recompute a full SVD-based plane fit using pixels whose depth lies within tolerance $\tau$ of the per-row model. The fitted plane $\mathbf{n}^\top \mathbf{p} = d$ is then used to gate candidate floor pixels. A pixel is classified as floor only if its residual to the plane is below 0.02\,m and its vertical gradient is below 0.04.

The near-field line mask is defined as:
\begin{equation}
  M_{\mathrm{near}} = \mathrm{medBlur}\!\left(M_{\mathrm{color}} \wedge
  M_{\mathrm{floor}}\right).
  \label{eq:mask}
\end{equation}
When $M_{\mathrm{near}}$ falls below the area threshold, the controller switches to a relaxed far-field mask with a fixed hue band (area $\ge 500$ px) and slows to 65\% of nominal speed to preserve heading while the near-field view remains partially occluded.

\subsubsection{Depth-Fused Obstacle Detector}
\label{subsec:obstacle_detect}

The \emph{Depth-Fused Obstacle Detector} (Figure~\ref{fig:framework}) detects obstacles within a forward image corridor spanning half the image width by fusing geometric and semantic cues.

\textbf{Geometric obstacles} are extracted from depth residuals. Points whose depth exceeds the per-row floor model by more than $k_{\mathrm{mad}} \cdot \mathrm{MAD} + \tau_{\mathrm{margin}}$ with MAD-based adaptive threshold $k_{\mathrm{mad}}=2$ and $\tau_{\mathrm{margin}}=0.05\,\mathrm{m}$ are treated as candidate obstacles. These candidates are further filtered using real-world size constraints, requiring a minimum width and height of $0.03\,\mathrm{m}$, a pixel area of at least $140\,\mathrm{px}$, and sufficient vertical gradient. To reduce incorrect detections, blobs must also persist for at least two consecutive frames.

\begin{figure}[t]
  \centering
  \includegraphics[width=\linewidth]{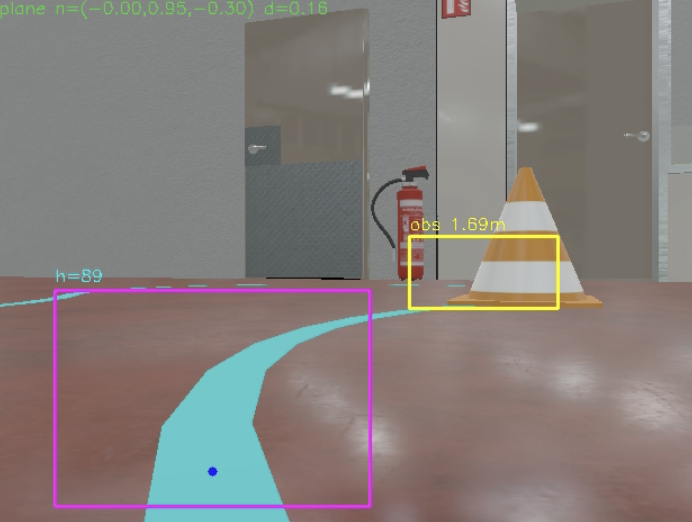}
  \caption{Simulation view with fused obstacle detection and a forward corridor.}
  \label{fig:obstacle}
\end{figure}

\textbf{Semantic obstacles} are detected using YOLOv8n. For each bounding box, the aligned depth image is used to compute a median depth estimate, and the resulting candidates are ranked by their distance within the forward corridor. The fused obstacle output is illustrated in Figure~\ref{fig:obstacle}, and the ranked obstacle list is forwarded to the Decision Layer.

\subsubsection{Lightweight Visual Odometry and Knowledge Base}
\label{subsec:vo}

The \emph{Lightweight Visual Odometry} module (Figure~\ref{fig:framework})
produces incremental pose estimates that populate the Knowledge Base breadcrumb
map.  The pipeline consists of: (i)~GFTT feature detection; (ii)~ORB descriptor
extraction; (iii)~KNN matching with Lowe's ratio test; and (iv)~ a five-point
essential matrix solver with RANSAC. These estimates are then linked to form an incremental robot trajectory.

Whenever the near-field line mask is active, and the robot has moved more than 0.10\,m since the last stored pose, the current trajectory pose is appended to a FIFO list (\verb|line_map|).
This list captures recently observed line locations and serves as the target set for Stage-2 recovery. Because the VO system is monocular and does not use loop closure, the breadcrumb trail gradually drifts over longer runs. This remains the main limitation of Stage-2 recovery and is discussed further in Section~\ref{subsec:failure}.

\subsection{Phase 3: Decision Layer (Plan)} \label{subsec:decision}

After state estimation, the Decision Layer determines the robot's action for each control tick by applying three mutually exclusive priorities (Phase~3, Figure~\ref{fig:framework}):
\begin{enumerate}
  \item[\textbf{P1.}] \textbf{Obstacle Avoidance}:  if the nearest fused obstacle lies within the warning threshold ($d \le d_{\mathrm{warn}}= 1.40$ m), the avoidance behavior takes precedence over all other actions.
  \item[\textbf{P2.}] \textbf{Line Following}: in the absence of an immediate obstacle threat, the robot continues tracking the guide line using PD control whenever either the near-field mask or the relaxed far-field mask is active.
        
  \item[\textbf{P3.}] \textbf{Recovery}: if neither line mask has been detected for longer than the 0.6\,s grace window and no obstacle is blocking, the two-stage recovery algorithm is engaged.
\end{enumerate}
All timing variables associated with these priorities are maintained in the Knowledge Base, allowing state to persist across ticks without external middleware. Consequently, the complete MAPE-K loop is executed within a single 50\,ms control tick, removing the need for a separate adaptation manager~\cite{white2004:architectural}.

\subsection{Phase 4: Actuation (Execute)} \label{subsec:execute}

In Phase~3, the chosen behavior calculates the wheel speeds, limits them to
safe hardware values, and sends them to the motors in Phase~4 (Figure~\ref{fig:framework}) for the execution.

\subsubsection{P1: Obstacle Avoidance}

When the Decision Layer selects P1, the closest obstacle in the fused list controls the avoidance response. If $d \le d_{\mathrm{warn}}=1.40$\,m, the robot slows and pivots toward the larger left-right depth gap. If $d \le d_{\mathrm{near}}=0.75$\,m, it carries out an immediate bounded pivot (hold 1.0--3.0\,s, hard timeout 3.0 s). Normal line tracking resumes once clearance is greater than 1.65\,m.

\subsubsection{P2: Line Following (PD Control)}

The robot follows the centroid $(c_x, c_y)$ of the largest connected component in $M_{\mathrm{near}}$ using a PD controller:
\begin{equation}
  e_k = \frac{c_x - w/2}{w/2},\quad
  u_k = K_p\,e_k + K_d\,\frac{e_k - e_{k-1}}{\Delta t},
  \label{eq:pid}
\end{equation}
with $K_p = 0.9$, $K_d = 0.1$, and $\Delta t = 0.05$\,s (20\,Hz).

Forward speed is decreased when lateral error becomes large:
\begin{equation}
  v_k = v_{\mathrm{base}} \cdot \max\!\left(v_{\mathrm{min}},\;
        1 - \beta\,|e_k|\right),
  \label{eq:speed_scaling}
\end{equation}
where $\beta$ determines how aggressively speed is reduced. Wheel velocities are given by:
\begin{equation}
  \omega_L = v_k + u_k,\quad \omega_R = v_k - u_k,
  \label{eq:wheel_cmd}
\end{equation}
and are clipped to the hardware limits before being sent to the robot. When the far-field mask is the only cue, $v_k$ is scaled by 0.65 and $u_k$ is derived from the far centroid, which provides degraded but stable heading maintenance.

\subsubsection{P3: Two-Stage Recovery Algorithm}
\label{subsec:recovery}

\begin{figure}[t]
  \centering
  \includegraphics[width=0.92\linewidth]{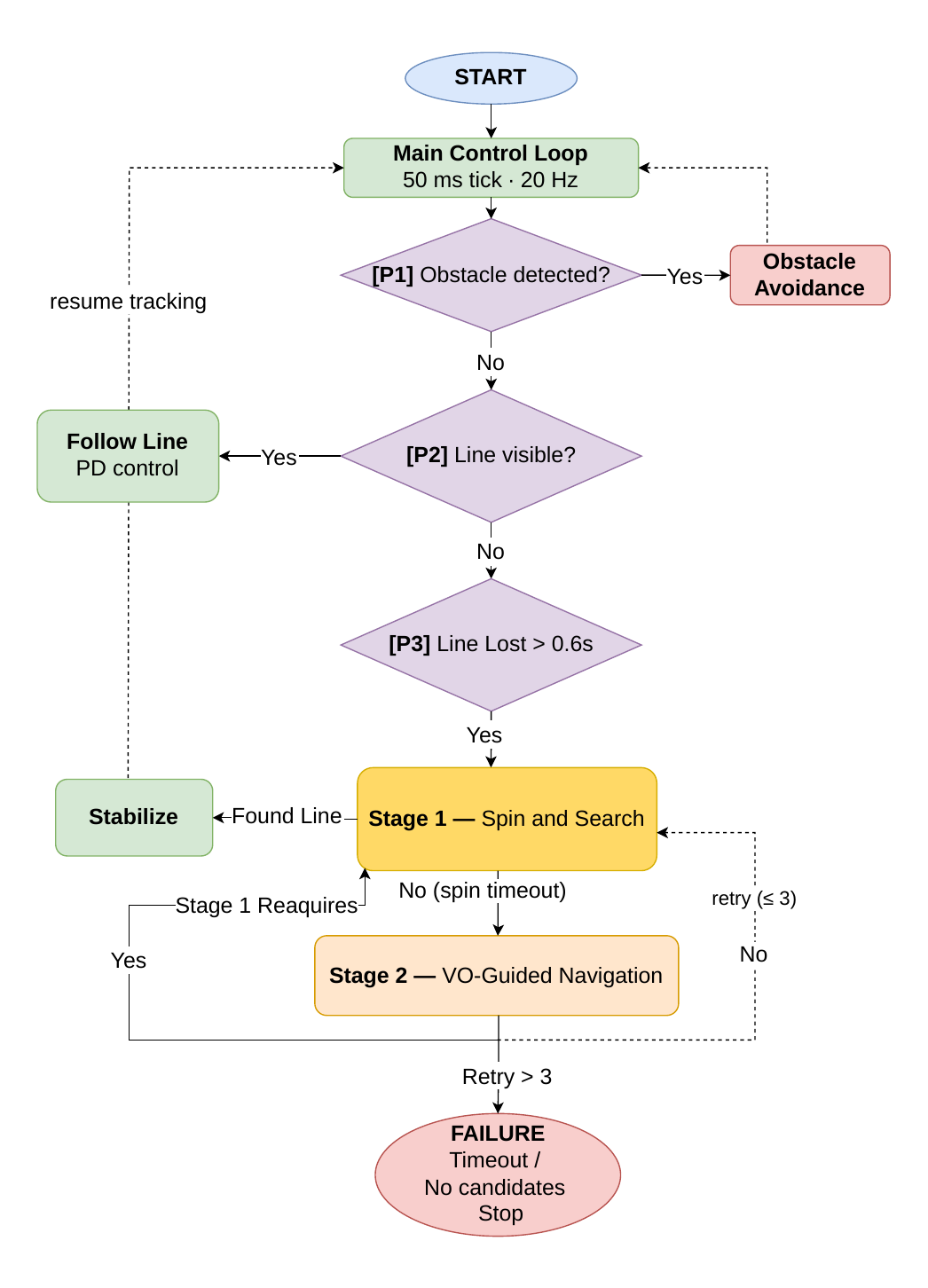}
  \caption{Two-stage recovery control flow (Phase~4, Figure~\ref{fig:framework}).
    Stage~1 (spin-and-search) attempts reacquisition first; Stage~2
    (VO-guided navigation to a Knowledge-Base breadcrumb) is escalated to
    only on Stage-1 failure.  Up to three spin--navigate cycles are permitted
    before a failure episode is declared}
  \label{fig:flowchart}
\end{figure}

When P3 is selected, the two-stage recovery algorithm is activated. The 0.6\,s grace window requires the line cues to remain absent continuously before recovery is triggered, so a brief dropout alone does not qualify. Figure~\ref{fig:flowchart} presents the complete recovery control flow within Phase~4 of Figure~\ref{fig:framework}.

\head{Stage 1: Spin-and-Search.}
Once recovery starts, $h^*$ is held at its latest value and the detection criteria are
relaxed: $\delta_h \mathrel{+}= 8$ hue units, floor-plane tolerance increases by
$0.02$ m, and the gradient ceiling is lifted to 0.04. The robot then rotates in place
for up to $T_{\mathrm{spin}}=2.0$ s at an angular fraction of 0.65. A detection is
accepted only after $N_{\mathrm{confirm}}=5$ consecutive frames satisfy both conditions:
(a) the near-field blob area is at least $900$ px; and (b) the blob centroid sits below
45\% of the image height, helping filter out ceiling reflections and partial views. After a
successful detection, the system pauses for 0.6 s to stabilize before resuming nominal PD
control, followed by a 0.8 s relaxed-threshold window to lower the chance of immediately
falling back into recovery.

\head{Stage 2: VO-Guided Navigation.}
If Stage~1 fails, the nearest breadcrumb pose more than 0.10\,m away is selected from \verb|line_map| in the Knowledge Base. A proportional angle-then-distance controller then guides the robot to that pose:
\begin{equation}
u_{\theta} = K_{p,\theta}\,(\theta_{\mathrm{tgt}} - \theta),\quad
u_d = K_{p,d}\,(d_{\mathrm{tgt}} - d),
\label{eq:nav}
\end{equation}
using an 8.0\,s timeout for each step. On arrival, Stage~1 is repeated. Up to $N_{\mathrm{max}}=3$ spin--navigate cycles are allowed; if the line still has not been reacquired after that, the episode is considered a failure.

Algorithm~\ref{alg:recovery} summarizes the Two-Stage Recovery Algorithm; the full set of controller parameters is listed in Table~\ref{tab:params} in Appendix~\ref{app:params}

\begin{algorithm}[t]
\small
\caption{Two-stage recovery (called each tick after grace window expires)}
\label{alg:recovery}
\begin{algorithmic}[1]
\Procedure{Recovery}{}
  \If{not \emph{line\_lost}}
    \State mark lost; freeze $h^*$; record $t_{\mathrm{start}}$; push VO pose
  \EndIf
  \If{\Call{SpinAndSearch}{}} \Comment{Stage 1}
    \If{\Call{Stabilize}{}} \State \textbf{return} \emph{success} \EndIf
  \EndIf
  \If{attempts $\ge N_{\mathrm{max}}$} \Comment{budget exhausted}
    \State stop; \textbf{return} \emph{failure}
  \EndIf
  \If{not \Call{NavigateToNearest}{}} \Comment{Stage 2}
    \State \textbf{return} \emph{failure}
  \EndIf
\EndProcedure
\Statex
\Procedure{SpinAndSearch}{}
  \State stability\_count $\leftarrow 0$;\; $t_0 \leftarrow$ now
  \While{now $-t_0 < T_{\mathrm{spin}}$}
    \State rotate in place with relaxed thresholds
    \If{line confirmed: area \& bottom-of-image check}
      \State stability\_count $\mathrel{+}=1$
      \If{stability\_count $\ge N_{\mathrm{confirm}}$}
        \State \textbf{return} \emph{true}
      \EndIf
    \Else \; stability\_count $\leftarrow 0$
    \EndIf
  \EndWhile
  \State attempts $\mathrel{+}=1$;\; \textbf{return} \emph{false}
\EndProcedure
\Statex
\Procedure{NavigateToNearest}{}
  \State pick pose $p^* \in$ \verb|line_map| with $\|p^*\!-\!p\| > 0.10$\,m
  \State drive with angle--distance control, timeout $= T_{\mathrm{step}}$
  \State \textbf{return} \emph{reached} or \emph{timeout}
\EndProcedure
\end{algorithmic}
\end{algorithm}

\section{Experimental Setup}
\label{sec:eval}

The simulation environment stresses the \emph{Depth-Gated Line Tracker} and \emph{Depth-Fused Obstacle Detector} (Phase~2); course geometry determines which branch of the Phase-3 \emph{Decision Layer} is activated most frequently; and fault injection directly triggers the \emph{Two-Stage Recovery Module} (Phase~4) in Figure~\ref{fig:framework}.

\subsection{Robot Platform}
\label{subsec:platform}

\begin{figure}[t]
  \centering
  \includegraphics[width = 0.45\textwidth]{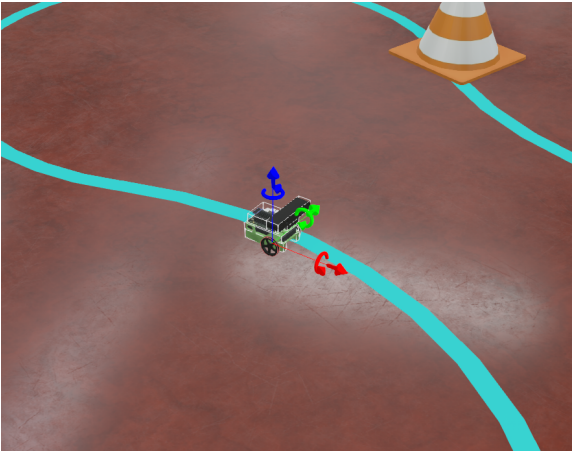}
  \caption{Top view of the JetBot differential-drive UGV used in all
    experiments. The compact chassis houses two independently driven wheels
    and a front-facing Orbbec Astra RGB-D camera, representing the minimum
    hardware configuration targeted by this work.}
  \label{fig:jetbot_exp}
\end{figure}

The target hardware is a JetBot-class differential-drive UGV representing the minimum viable configuration considered in this work: two independently driven wheels, an onboard CPU, and a single front-facing camera. The JetBot platform, originally developed by NVIDIA as an accessible research robot, has been widely adopted for embedded robotics owing to its compact form factor and modest power budget~\cite{2026:nvidiaaiiot,:ai}. These properties make it representative of cost-sensitive deployments in logistics, inspection, and small-scale agriculture, where rich sensor suites are not economically justifiable. A top-down view of the platform is shown in Figure~\ref{fig:jetbot_exp}.

The depth-sensing modality is the Orbbec Astra camera introduced in Section~\ref{subsec:hardware}, which pairs its RGB sensor with a structured-light depth unit~\cite{:astra}. Its effective indoor depth range of 0.6--8\,m covers both the obstacle warning zone ($d_{\mathrm{warn}}=1.40$\,m) and the near-field line-tracking region, so no separate laser or ultrasonic sensor is required.

Camera intrinsic parameters are derived analytically from the reported field of view rather than from a physical calibration target, since the Webots simulator provides consistent sensor geometry. The horizontal focal length is approximated as
\begin{equation}
  f_x = f_y = \frac{w}{2\,\tan(\theta_{\mathrm{HFoV}}/2)},
  \label{eq:focal}
\end{equation}
where $w = 640$\,px and $\theta_{\mathrm{HFoV}} \approx 58^{\circ}$, giving $f_x \approx 617$\,px. The principal point is set to the image center $(c_x, c_y) = (320, 240)$. These intrinsics are used by the VO module to back-project depth pixels into 3D and to estimate the essential matrix from matched feature correspondences.

\subsection{Simulation Environment}

\begin{figure*}[t]
  \centering
  \includegraphics[width = 0.8\textwidth,height=0.40\textheight]{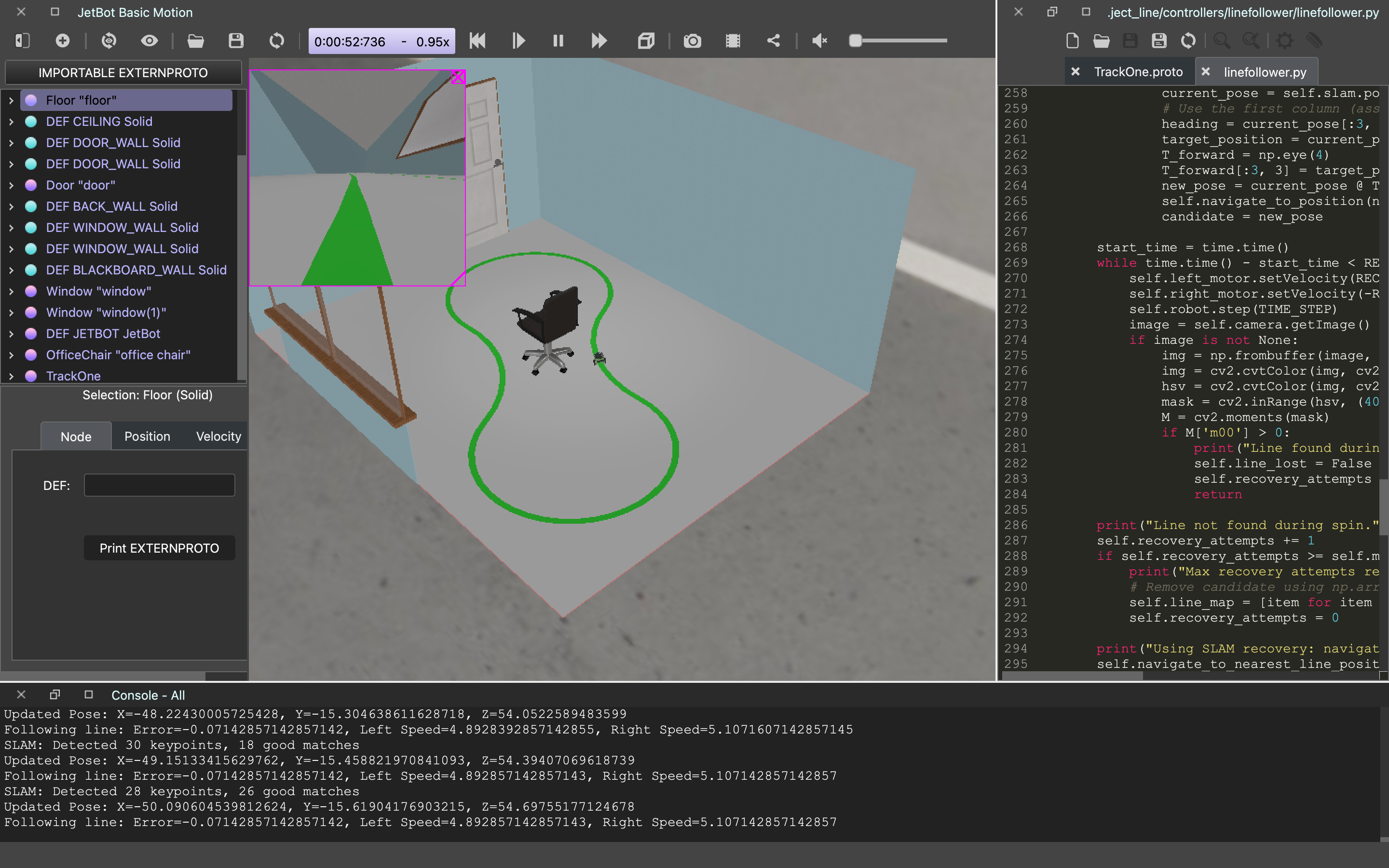}
  \caption{Webots simulation interface used in all experiments. The 3D scene
    editor (centre) displays the robot and environment model; the side panels
    provide real-time RGB and depth sensor feeds used during algorithm
    development and fault-injection testing.}
  \label{fig:webots_view}
\end{figure*}

All experiments were carried out in Webots~\cite{:cyberbotics}, an
open-source robot simulator that uses the Open Dynamics Engine (ODE) for
rigid-body physics and offers realistic RGB and depth sensor models. Webots
was preferred over alternatives such as Gazebo because it has lower CPU overhead,
better documentation, and a more integrated Python API for rapid
iteration~\cite{ayala2020:comparison}. Figure~\ref{fig:webots_view} shows the Webots interface used throughout all experiments, including the 3D scene editor, the robot model, and the sensor visualization panels that were used to monitor RGB and depth streams during development.

\begin{figure}[t]%
  \centering\tableheadfont%
  \subfloat[\tableheadfont Office world (baseline).]{
    \includegraphics[width=.395\linewidth]{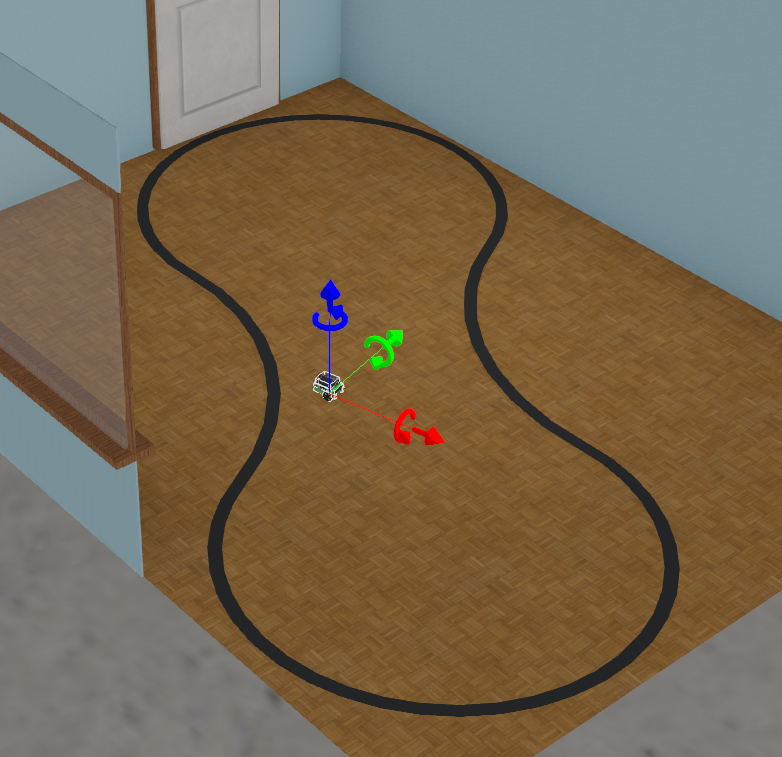}}%
  \hspace*{2pt}%
  \subfloat[\tableheadfont Factory hall world.]{%
    \includegraphics[width=.54\linewidth]{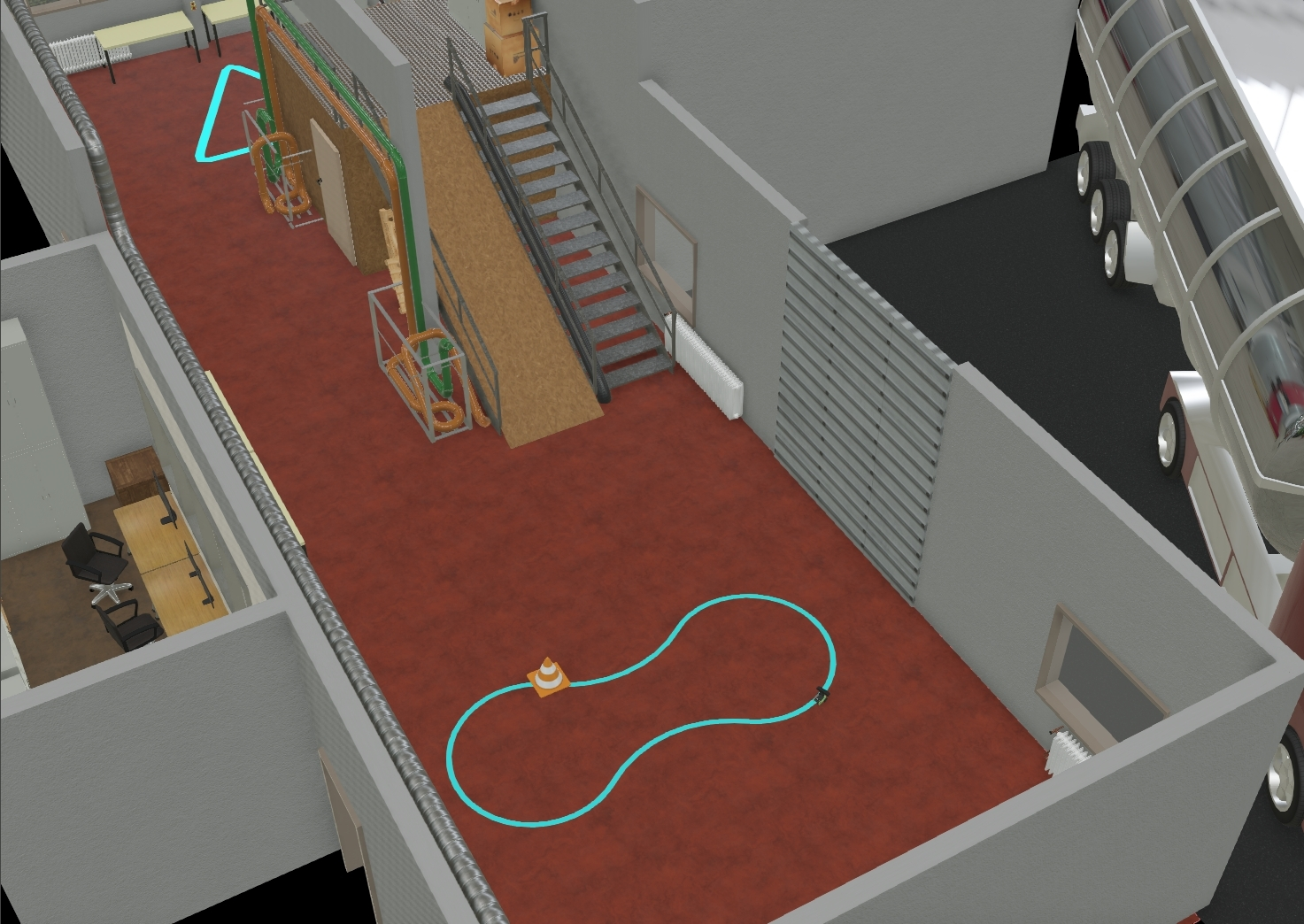}}%
  \caption{Simulated environments. The factory hall includes reflective flooring,
    shelving, and ramps that challenge the Phase-2 perception modules.}
  \label{fig:worlds}\vspace*{-3ex}
\end{figure}

The approach is evaluated on two world types constructed in the Webots simulator (Figure~\ref{fig:worlds}):
\begin{itemize}
  \item \textbf{Factory hall}: a long-corridor setting with shelving, ramps, and a reflective floor. The reflective surface challenges the SVD plane gating in the \emph{Depth-Gated Line Tracker}, while the clutter frequently triggers the \emph{Depth-Fused Obstacle Detector}.
\item \textbf{Office}: a simpler, structured layout used for baseline calibration of the Phase-2 detection thresholds.
\end{itemize}

\subsection{Line Courses}

\begin{figure}[t]%
  \centering\tableheadfont%
  \subfloat[\centering\tableheadfont Course 1: \newline Smooth oval.]{%
    \includegraphics[width=0.32\linewidth]{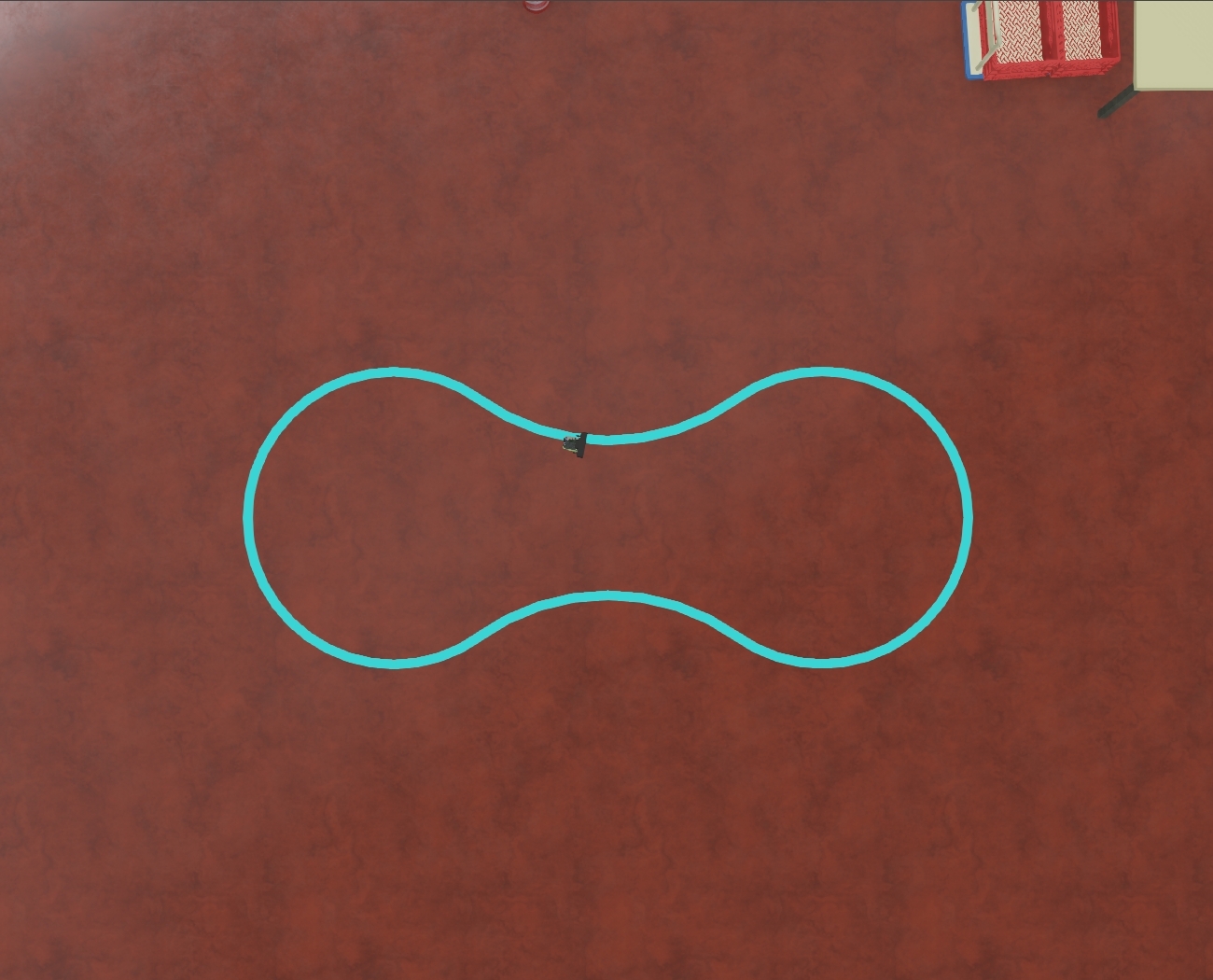}}~
  \subfloat[\centering\tableheadfont Course 2: \newline Arrow turns.]{%
    \includegraphics[width=0.32\linewidth]{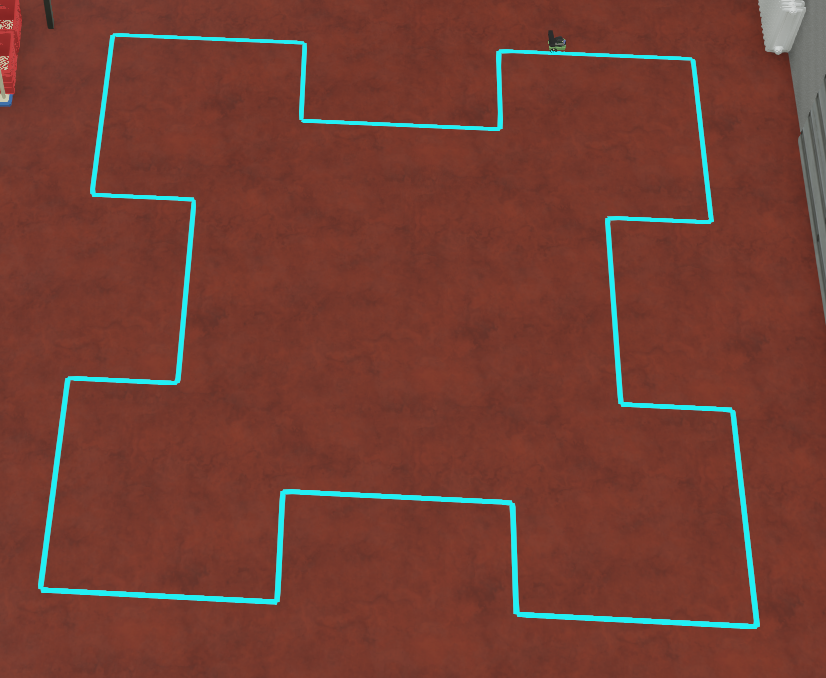}}~
  \subfloat[\centering\tableheadfont Course 3: \newline Acute triangle.]{%
    \includegraphics[width=0.335\linewidth]{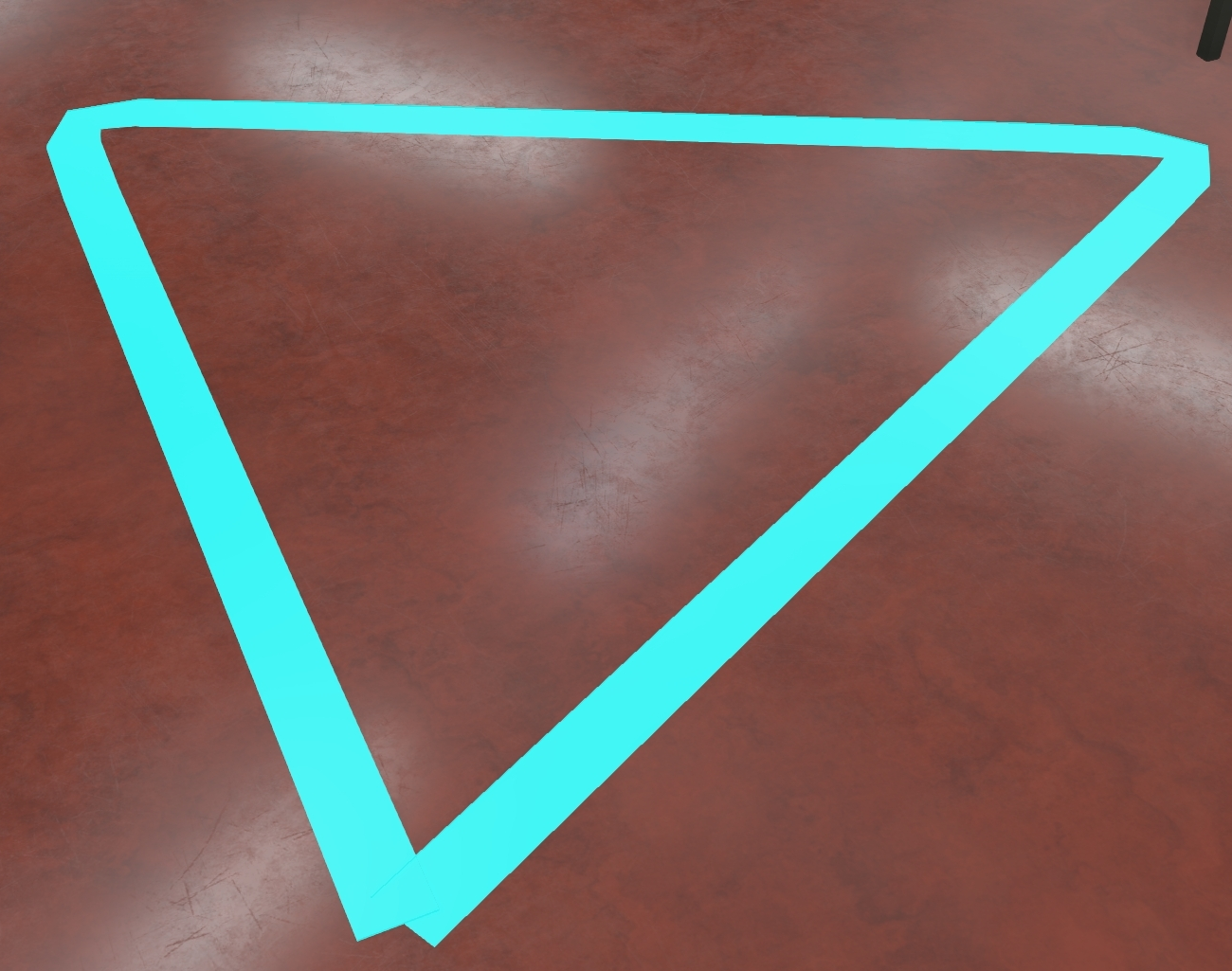}}
  \caption{The three line courses used in our evaluation. The course shape determines
    the line-loss frequency and which stage of the Phase-4 Recovery Module
    dominates.}
  \label{fig:courses}
\end{figure}

Three guideline courses with progressively greater geometric complexity were painted onto the floors of the simulation worlds (Figure~\ref{fig:courses}). Course geometry largely determines which stage of the Phase-4 \emph{Recovery Module} is triggered most often.
\begin{itemize}

\item \textbf{Course~1 (oval)}: A smooth, continuously curving route with no sharp corners. When the line is lost, it falls well outside the camera FoV, so the Phase-3 Decision Layer typically escalates to P3 and invokes \emph{VO-Guided Navigation} (Stage~2).
\item \textbf{Course~2 (narrow turns)}: An indented layout with short straight segments and abrupt direction changes. Line loss often still leaves part of the line visible, so \emph{Spin-and-Search} (Stage~1) resolves most episodes within the 2 s spin budget.
\item \textbf{Course~3 (acute-corner triangle)}: A triangular route with three acute corners that causes immediate, complete line loss. Stage~2 is needed frequently, but the compact geometry keeps Knowledge-Base breadcrumbs within short travel distances.
\end{itemize}

\subsection{Fault Injection Protocol}

Line-loss episodes were induced by programmatically removing the painted line
texture from the camera view during operation, simulating real-world cases such
as worn floor markings, strong glare, or sudden heading deviation after obstacle
avoidance. Boxes and cones were also placed near or on the line to
independently stress the avoidance module.

Three fault categories were evaluated in the study:
\begin{enumerate}
\item \textbf{Visual loss}: the guideline texture was temporarily removed from the camera view during operation. The near-field mask vanished immediately, while far-field cues could persist briefly when the robot remained approximately aligned with the course. This constituted the primary fault mode and accounted for all 119 evaluation episodes.
\item \textbf{Dynamic obstacles}: boxes or cones were introduced on or near the guide line. These disturbances activated the avoidance module and could induce secondary line loss if the resulting avoidance maneuver moved the line outside the camera field of view.
\item \textbf{Pose displacement}: the robot was teleported between frames to assess Stage-2 navigation under large initial pose errors. This condition was used only during preliminary calibration and was not included in the 119-episode evaluation set.
\end{enumerate}
Experiments consisted of multiple standardized laps per course starting from
fixed initial poses.  An episode begins when the controller first detects line
loss (grace window starts) and ends when either: (a)~$N_{\mathrm{confirm}}=5$
consecutive frames confirm stable reacquisition; or (b)~all recovery cycles are
exhausted.  Timestamps at both events are logged to compute TTR.  Only visual-loss
(type~1) episodes are included in the reported statistics.

\subsection{Research Questions}

Three research questions frame the evaluation:
\begin{enumerate}[label=\textbf{RQ\arabic*}, leftmargin=*]
  \item \emph{What autonomous recovery strategies can enable a vision-based UGV to regain operational status after losing its navigation cue?}
  \item \emph{What are the design trade-offs between achieving high recovery resilience and maintaining low hardware cost in camera-first UGV systems?}
  \item  \emph{To what extent can simulation-based fault injection be used to evaluate and refine a recovery algorithm for low-cost UGVs?}
\end{enumerate}

\subsection{Evaluation Metrics}\label{sec:metrics}

Each episode is evaluated using three metrics:
\begin{enumerate}
  \item \textbf{Recovery success rate}: the share of episodes in which line reacquisition is successful. We report 95\% Wilson score confidence intervals to account for finite sample sizes.
  \item \textbf{Time to Recovery (TTR)}: the elapsed time from fault onset to confirmed reacquisition in successful episodes. Results are reported as mean~$\pm$~SD, median, and the p10--p90 range.
  \item \textbf{Recovery mechanism}: an episode-level classification of Stage-1 (spin-only) versus Stage-2 (VO-navigation), highlighting how course geometry influences the algorithm's behavior.
\end{enumerate}

\section{Results and Discussion}
\label{sec:results}

A detailed analysis of the results is presented in this section. Table~\ref{tab:results} and Figures~\ref{fig:mechanisms} and ~\ref{fig:ttr_hists} summarize the per-course results, while Table~\ref{tab:comparison} provides cross-system context. Unlike the qualitative positioning in Table~\ref{tab:positioning}, this table reports quantitative recovery times, which are limited to the prior camera-only system that reports a directly comparable metric~\cite{lee2012:visionbased}. The SLAM and deep-learning systems in Table~\ref{tab:positioning} do not report a line-loss recovery time and are therefore not included here.

\subsection{RQ1: Effectiveness of Two-Stage Recovery}

\subsubsection{Overall Success and Speed}

Across 119 induced line-loss episodes, the proposed system successfully recovered in \textbf{103 cases (86.6\%)}. The corresponding 95\% Wilson confidence interval was $[0.79,\;0.92]$ (Table~\ref{tab:results}), the lower CI bound of 0.79 indicates the result is unlikely to be a sampling artifact. The overall median Time to Recovery (TTR) was \textbf{3.26\,s}, with a mean of 3.48 s $\pm$ 2.55 s. The broad spread in recovery times (p10 = 0.41\,s, p90 = 7.27\,s) reflects the two operating modes of the staged strategy, with rapid spin-based reacquisition in some cases and slower VO-guided navigation in others.

\begin{table*}[t]
\centering
\caption{Per-course recovery results. ``Spin'' = Stage-1 spin-only resolutions;
``Nav'' = Stage-2 VO-navigation resolutions. TTR statistics over successful
episodes only. (Evidence for RQ1 and RQ3.)}
\label{tab:results}
\begin{tabular*}{\textwidth}{@{\extracolsep{\fill}}lrrrccccc}
\toprule
 & \thead{Episode} & \thead{Success} & \thead{\%} &
 \thead{Mean$\pm$SD\,(s)} & \thead{Median\,(s)} & \thead{p10\,(s)} & \thead{p90\,(s)} & \thead{Spin~/~Nav} \\
\midrule
Course 1 & 26 & 20 & 76.9 & 3.99$\pm$1.92 & 3.48 & 2.64 & 7.34 &  2~/~18 \\
Course 2 & 49 & 45 & 91.8 & 2.51$\pm$2.57 & 0.85 & 0.33 & 6.93 & 24~/~21 \\
Course 3 & 44 & 38 & 86.4 & 4.36$\pm$2.42 & 4.27 & 1.21 & 8.33 &  8~/~30 \\
\midrule
Overall  & 119 & 103 & 86.6 & 3.48$\pm$2.55 & 3.26 & 0.41 & 7.27 & 34~/~69 \\
\bottomrule
\end{tabular*}
\end{table*}

\subsubsection{Stage-1 Spin-And-Search}

Stage-1 contributed 34 of the 103 successful recoveries (33\%) and typically completed reacquisition in under 1\,s. It has shown most effective results on Course~2, where the narrow-turn geometry often left part of the line visible after a heading deviation; as a result, 24 of 45 Course~2 successes (53\%) were resolved through spin-only recovery (Figure~\ref{fig:mechanisms}b, Table~\ref{tab:results}). The pronounced sub-1\,s peak in Figure~\ref{fig:ttr_c2} provides clear evidence that Stage-1 behaved as intended, delivering rapid recovery while the relaxed thresholds and 5-frame confirmation rule helped reduce false positives.

\subsubsection{Stage-2 VO-Guided Navigation}

\begin{figure*}[t]%
  \centering\tableheadfont%
  \subfloat[\tableheadfont Course 1: Spin~2, Nav~18.]{%
    \includegraphics[width=0.34\linewidth, trim=10pt 5pt 0pt 27pt, clip]{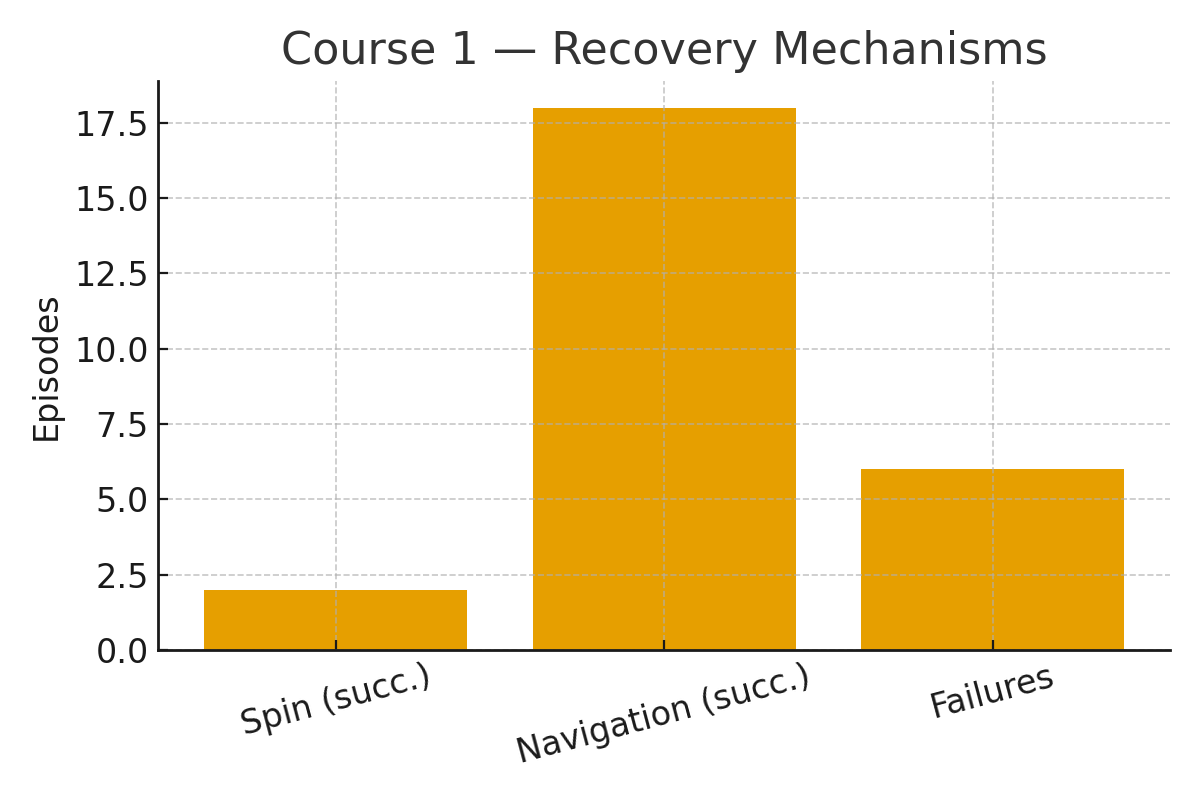}}%
  \subfloat[\tableheadfont Course 2: Spin~24, Nav~21.]{%
    \includegraphics[width=0.33\linewidth, trim=25pt 5pt 0pt 27pt, clip]{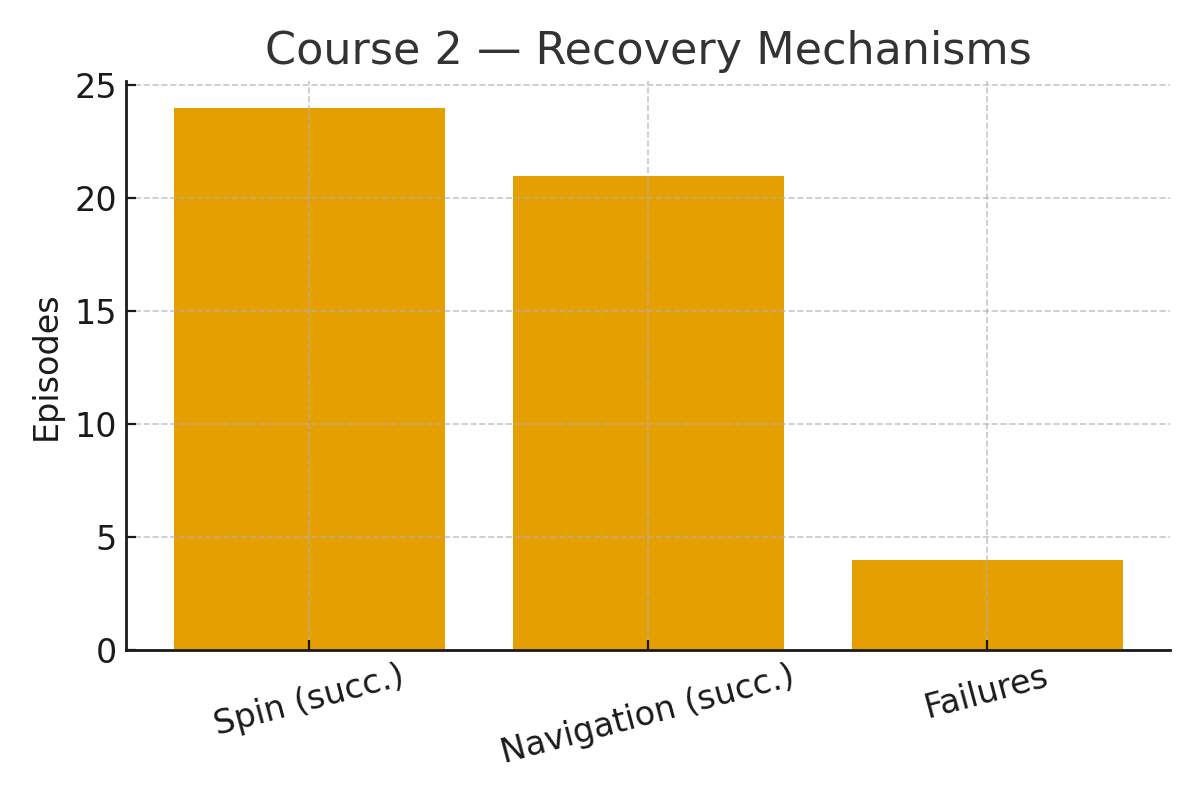}}%
  \subfloat[\tableheadfont Course 3: Spin~8, Nav~30.]{%
    \includegraphics[width=0.33\linewidth, trim=25pt 5pt 0pt 27pt, clip]{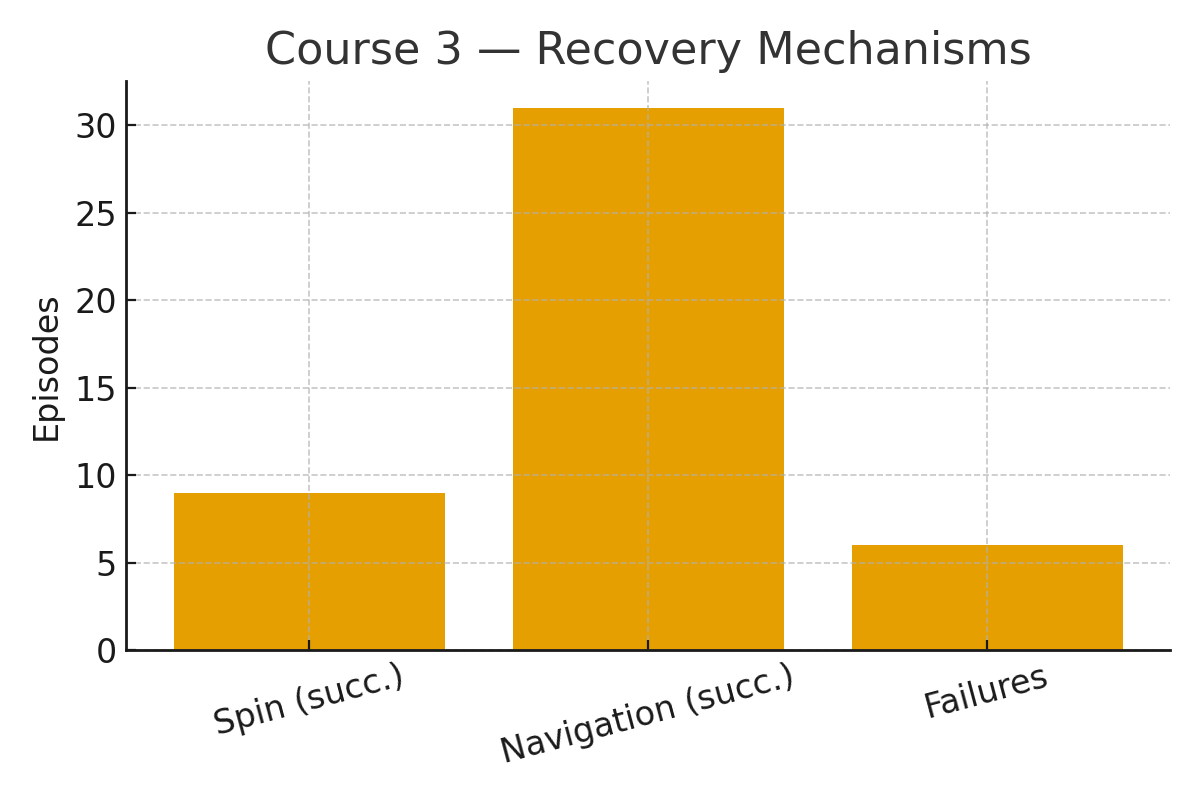}}%
  \caption{Recovery mechanism counts per course (RQ1). Stage-1 spin dominates
    where the line stays partially visible (Course~2); Stage-2 VO navigation
    dominates where it disappears completely (Courses~1 and~3). The split
    validates the staged escalation design.}
  \label{fig:mechanisms}
\end{figure*}

\begin{figure*}[t]
  \centering\tableheadfont%
  \subfloat[\tableheadfont Course~1: Time to Recovery (TTR).]{%
    \includegraphics[width=0.34\linewidth, trim=10pt 5pt 0pt 27pt, clip]{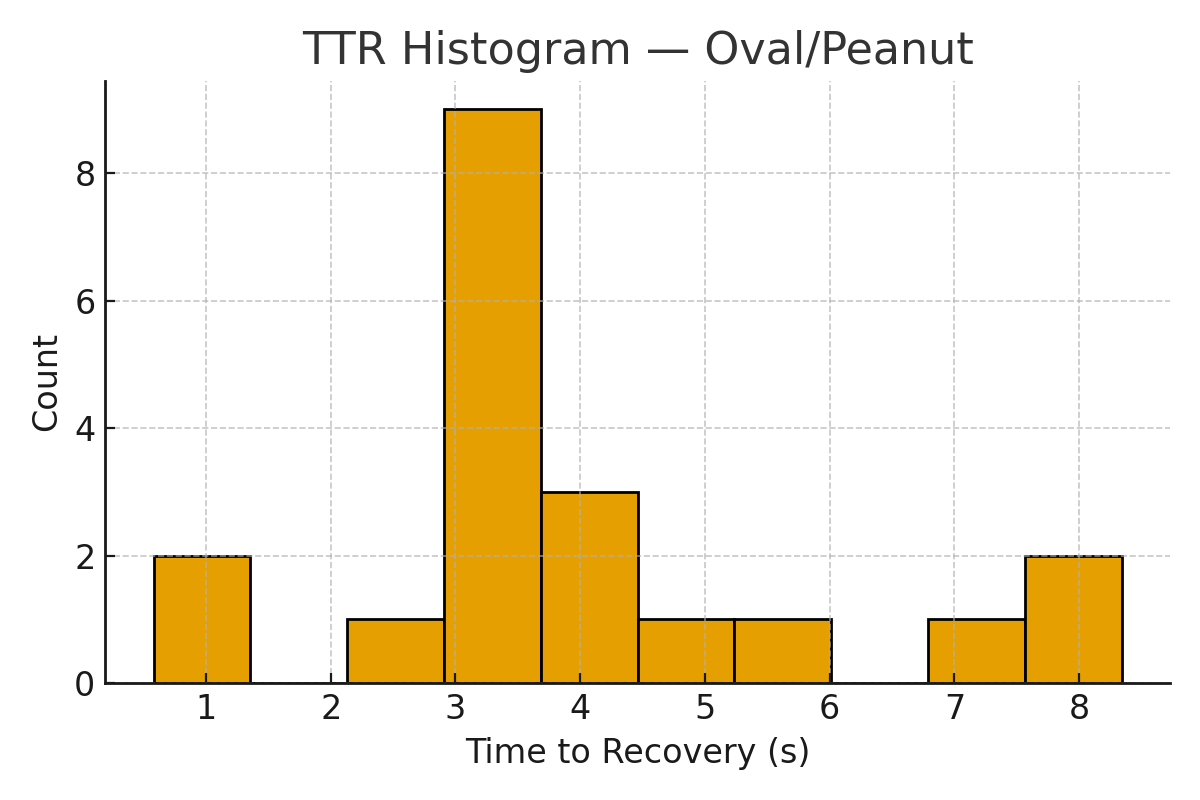}%
    \label{fig:ttr_c1}}%
  \subfloat[\tableheadfont Course~2: Time to Recovery (TTR).]{%
    \includegraphics[width=0.33\linewidth, trim=25pt 5pt 0pt 27pt, clip]{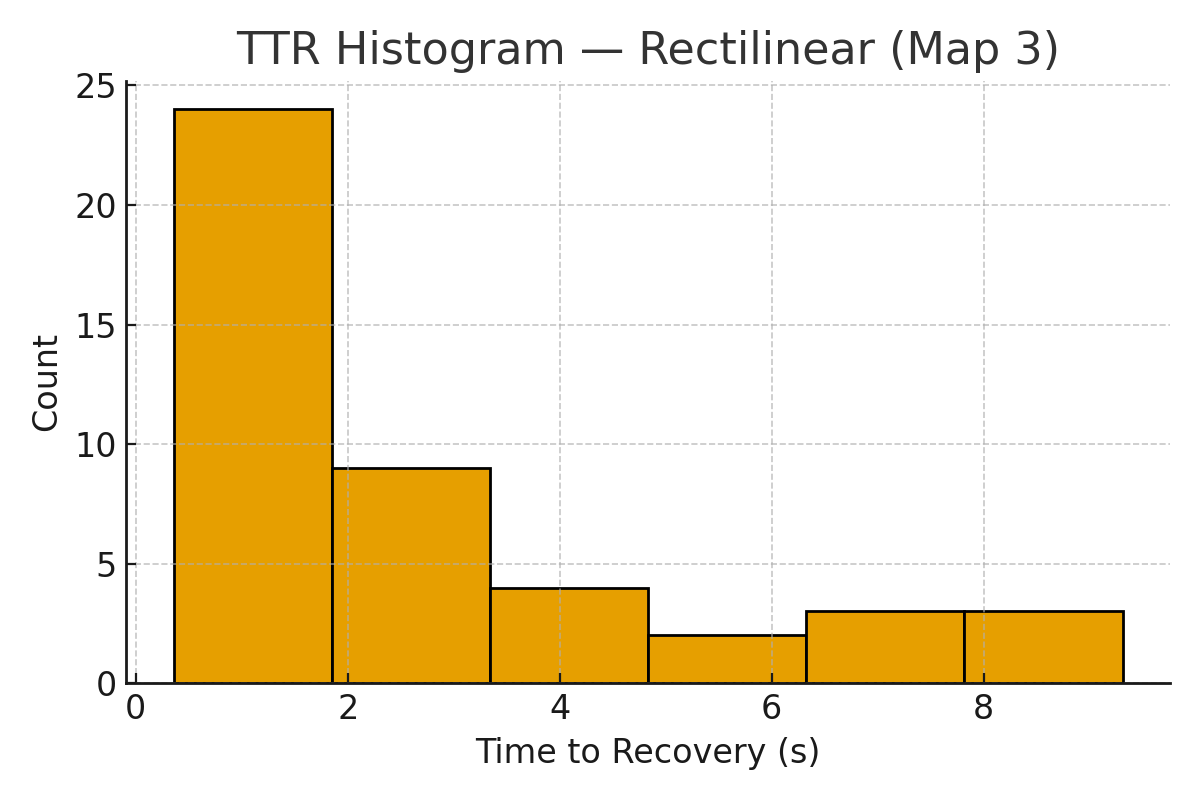}%
    \label{fig:ttr_c2}}%
  \subfloat[\tableheadfont Course~3: Time to Recovery (TTR).]{%
    \includegraphics[width=0.33\linewidth, trim=25pt 5pt 0pt 27pt, clip]{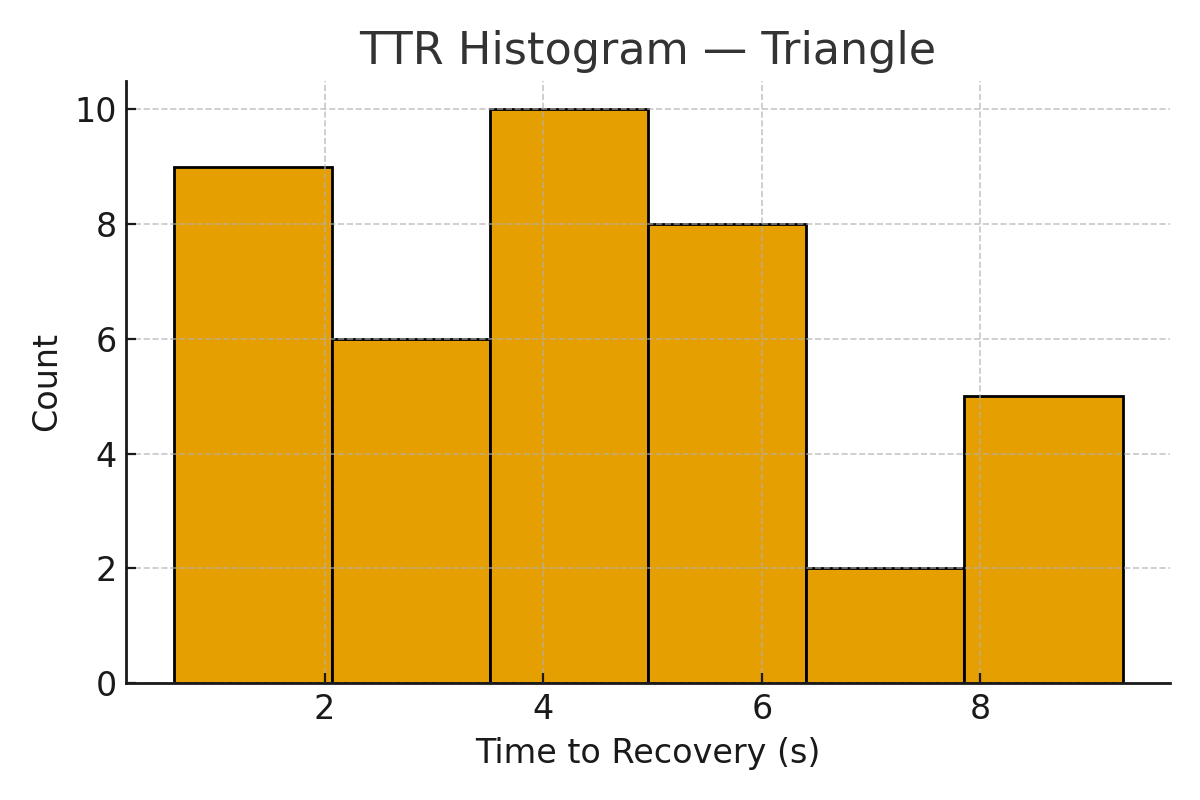}%
    \label{fig:ttr_c3}}%
  \caption{Per-course TTR histograms for successful episodes (RQ1). Course~2
    shows a clear bimodal distribution: a sub-1\,s peak (Stage-1 spin) and a
    2--7\,s tail (Stage-2 navigation). Courses~1 and~3 are dominated by Stage-2,
    producing unimodal distributions shifted right. The shape of each
    histogram directly reflects the mechanism counts in Figure~\ref{fig:mechanisms}.}
  \label{fig:ttr_hists}
\end{figure*}

When spin-only recovery was insufficient, Stage~2 guided the robot to the nearest stored VO breadcrumb before the recovery sequence was reinitiated. Stage~2 accounted for 69 of the 103 successful recoveries (67\%) and was particularly effective under complete line-loss conditions, contributing to 90\% of Course~1 successes (18/20, Figure~\ref{fig:mechanisms}a) and 79\% of Course~3 successes (30/38, Figure~\ref{fig:mechanisms}c). Stage~2 recovery times ranged from 2 to 9\,s, depending on breadcrumb proximity, which produced the pronounced right tail observed in all three histograms (Figure~\ref{fig:ttr_hists}). Although Stage~2 was used frequently, Course~3 achieved a median TTR of 4.27\,s because of its shorter inter-corner breadcrumb spacing, whereas Course~1 required longer navigation segments, resulting in a median TTR of 3.48\,s and the lowest overall success rate (76.9\%, Table~\ref{tab:results}).

\subsubsection{Post-Recovery Stability}

A 0.6\,s stabilization pause, followed by a 0.8\,s relaxed-threshold window after reacquisition, prevented oscillatory re-entry into recovery in almost all episodes. Only a small number of back-to-back activations were observed, and these occurred only on Course~3, where recovery often ended near an acute corner. Overall, the robot resumed autonomous lap completion in the vast majority of episodes without operator assistance, demonstrating a high task-continuity rate in the sense of~\cite{romero-garces2022:managing}.

\subsubsection{Failure Boundary}
\label{subsec:failure}

The 16 failed episodes, distributed as 6 on Course~1, 4 on Course~2, and 6 on Course~3, defined the conditions under which the strategy was not fully effective. Three failure modes were identified across these episodes: (i) \textbf{sparse breadcrumb coverage}, where obstacle avoidance displaced the robot beyond the reach of any stored pose and repeated spin attempts exhausted the recovery budget; (ii) \textbf{persistent occlusion}, where dynamic obstacles blocked the line throughout all recovery attempts and prevented both spin-based search and post-navigation recovery from succeeding; and (iii) \textbf{VO drift}, where monocular visual odometry drift during pure rotation or low-texture traversal caused Stage~2 to navigate toward an incorrect target, moving the robot farther from the line. Because individual episodes often combined more than one of these effects, we characterize the failure modes qualitatively rather than assigning a single cause to each of the 16 episodes. Each mode has a corresponding mitigation in Section~\ref{sec:future}.

Stage~1 spin-based recovery handled partial-visibility failures in under 1\,s (Figure~\ref{fig:ttr_c2}), whereas Stage~2 VO-guided navigation handled complete-loss scenarios in 2--9\,s (Figs.~\ref{fig:mechanisms}, \ref{fig:ttr_c1}, \ref{fig:ttr_c3}). The post-recovery grace window was necessary to prevent oscillatory re-entry into recovery. Overall, the strategy's effectiveness was primarily limited by breadcrumb coverage, persistent obstacle occlusion, and VO drift.

\takeaway{Answer to RQ1} {The two-stage recovery restored guideline tracking in 86.6\% of 119 line-loss episodes (95\% CI 0.79--0.92) at a median 3.26\,s, showing that camera-only, CPU-only recovery is effective across varied course geometries. }
\subsection{RQ2: Camera-First Cost-Resilience Trade-offs}

\subsubsection{Accuracy at Reduced Cost}

Table~\ref{tab:comparison} compared the proposed system with two published approaches in terms of recovery accuracy, mean TTR, GPU dependency, and sensor requirements. The proposed system achieved a recovery accuracy of \textbf{0.87 (103/119)}, comparable to the 87\% qualitative success rate reported by Lee~et~al.~\cite{lee2012:visionbased}. Since these studies used different platforms, sensors, and test conditions, this comparison indicates relative performance rather than a direct benchmark. Notably, this performance was obtained using only an Orbbec Astra RGB-D camera, with an estimated retail cost of \$100--150, and without relying on LiDAR, RADAR, GPS, or a GPU. In contrast, Lee~et~al.\ required additional sensors on an embedded ARM11 platform. These findings indicate that comparable recovery performance was achieved with a significantly simpler and lower-cost hardware configuration, enabled by VO-based breadcrumb navigation and a depth-gated line tracker.

\begin{table}[t]
\centering
\caption{Comparison with prior vision-based recovery systems (RQ2). Platform
and protocol differences preclude direct benchmarking; results are indicative
of relative performance across hardware tiers.}
\label{tab:comparison}\vspace*{-1ex}
\begin{tabular}{lc@{~}c}
\toprule
\thead[l]{Metric} & \thead{Proposed} & \thead{\cite{lee2012:visionbased}} \\
\midrule
Accuracy (succ./att.) & 0.87 (103/119) & 0.87$^a$          \\
Mean TTR (s)          & 3.48           & $\approx$3.23$^b$ \\
GPU required          & No             & No               \\
External sensors      & RGB-D only     & RGB + extras     \\
\bottomrule
\multicolumn{3}{l}{$^a$ Qualitative success rate; not per-episode.}\\
\multicolumn{3}{l}{$^b$ Global relocalization on embedded ARM11.}\\
\end{tabular}
\end{table}

\subsubsection{The Residual TTR Cost and Its Geometry-Dependence}

The mean Time to Recovery (TTR) of 3.48\,s is slightly higher than Lee~et~al.'s reported $\approx$3.23\,s (Table~\ref{tab:comparison}). Rather than reflecting a general slowdown, this difference is driven by Stage-2 navigation in geometrically demanding courses, where the camera-only setup is most constrained. On Course~1, long arcs move the line completely out of the field of view, so the median TTR remains 3.48\,s and Stage-2 accounts for 90\% of successful recoveries (Table~\ref{tab:results}, Figure~\ref{fig:mechanisms}a, Figure~\ref{fig:ttr_c1}). By contrast, on Course~2, the narrow-turn geometry preserves partial line visibility, reducing the median TTR to \textbf{0.85\,s} and allowing Stage-1 spin recovery to resolve 53\% of its successfull recoveries (Table~\ref{tab:results}, Figure~\ref{fig:mechanisms}b, Figure~\ref{fig:ttr_c2}), which is competitive with or faster than heavier sensor-equipped systems. Overall, the performance difference is \emph{geometry-dependent and predictable} rather than a uniform penalty.

\subsubsection{Cost of the Camera-Only Constraint in Failure Modes}

Two of the three failure modes identified under RQ1 arise directly from the camera-only, CPU-only design, reflecting the trade-off required to achieve self-healing under strict resource constraints. Sparse breadcrumb coverage could be addressed through global LiDAR-based localization, which enables relocalization without dependence on stored visual poses. VO drift remains an inherent limitation of monocular, depth-free odometry; a sensor-rich system could reduce this limitation through IMU fusion or loop closure. Collectively, these failures define the conditions under which the embedded MAPE-K loop cannot recover autonomously, and Section~\ref{sec:future} outlines concrete extensions to the knowledge base and recovery policy.

The camera-first design achieved recovery accuracy comparable to that of heavier prior systems (0.87 vs. 0.87, Table~\ref{tab:comparison}) while requiring substantially less hardware. 
The principal trade-off was a modest increase in mean TTR (3.48\,s vs. $\approx$3.23\,s), and this difference was strongly dependent on scene geometry. In environments that preserved partial line visibility, such as Course~2, the penalty was minimal, whereas fully occluded environments exposed the limitations of monocular VO. 
\takeaway{Answer to RQ2} {Camera-first, CPU-only recovery matched the 0.87 accuracy of a heavier prior system at a fraction of the hardware cost, with only a modest, geometry-dependent mean TTR increase (3.48\,s vs.\ $\approx$3.23\,s) and no unrecoverable failures. }
\subsection{RQ3: Simulation for Evaluation and Refinement}

\subsubsection{Scale and Precision of Fault Injection}

The Webots simulation enabled the safe, deterministic, and reproducible induction of \textbf{119 controlled line-loss episodes}, which would have been impractical on physical hardware because of cost, time, and safety constraints. Because fault onset was programmatically triggered and timestamped, exact TTR values were recorded for every episode (Table~\ref{tab:results}, p10--p90 columns). In contrast, physical hardware does not allow fault onset to be measured precisely, so TTR values would be approximate and the p10--p90 distribution in Table~\ref{tab:results} could not be reported reliably.

\subsubsection{Geometry-Driven Insight Through Course Variation}

The three courses were designed to evaluate how course layout influences the recovery strategy, and the results show that geometry is the main determinant of which recovery mechanism is selected. Course~2, with its narrow-turn configuration, achieved 53\% spin-only recoveries (Figure~\ref{fig:mechanisms}b), while the smoother oval of Course~1 achieved only 10\% spin-only recoveries (2/20; Figure~\ref{fig:mechanisms}a). This 43-percentage-point gap, combined with the consistent behavior observed within each course, indicates that recovery mode is strongly driven by geometry rather than random variation. Because the simulation provides precise episode-level labels, this pattern can be identified clearly; in a real-world setting, it would likely be less apparent due to confounding effects such as floor reflections, lighting variation, and sensor calibration drift.

\subsubsection{Algorithm Refinement Through Rapid Iteration}

Simulation enabled systematic tuning of the key thresholds, including hue tolerance, floor-plane tolerance, confirmation frame count, spin duration, and obstacle-distance parameters, through repeated fault injection without physical wear, battery constraints, or safety risk. %
For example, the 5-frame confirmation threshold ($N_{confirm}$) and the 0.6 s post-recovery stabilization window were set to their final values during repeated fault-injection runs in simulation. Their defaults are listed in Section~\ref{sec:eval}; a systematic ablation of these parameters is left to future work. Validating these refinements on physical hardware would have required substantially more experiments.

Webots-based fault injection produced 119 precisely timestamped episodes across three geometrically diverse courses, enabling exact Time to Recovery (TTR) distributions (Table~\ref{tab:results}, Figure~\ref{fig:ttr_hists}) and mechanism-level breakdowns (Figure~\ref{fig:mechanisms}) that would be difficult to obtain on physical hardware. This controlled setting isolated the geometry-driven split in recovery mechanisms and allowed rapid, safe refinement of the controller thresholds. %

\takeaway{Answer to RQ3} {Simulated fault injection produced 119 controlled, timestamped episodes that enabled exact TTR distributions and safe threshold tuning beyond what physical trials allow, while not replacing real-world validation.}

\subsection{Cross-Course Analysis}%

The three courses collectively illustrate a consistent relationship between course geometry and recovery mechanism. A comparison of mechanism distributions across courses (Figure~\ref{fig:mechanisms}) reveals a 43-percentage-point gap in Stage-1 spin recovery rates between the most favorable course (Course~2: 53\%) and the least favorable (Course~1: 10\%). This gap is not attributable to random variation but to systematic differences in how the three course layouts preserve or remove line visibility after a heading deviation.

From the perspective of the MAPE-K loop, this geometry-dependence reflects the interplay between the Monitor and Plan stages. On smooth oval courses (Course~1), continuous arcs allow the line to exit the camera field of view entirely, so the Monitor stage cannot supply any positive evidence during a spin, and the Plan stage must escalate to Stage-2 navigation. On narrow-turn courses (Course~2), abrupt direction changes produce line-loss episodes where the line remains partially visible from slightly different headings, allowing Stage-1 to confirm reacquisition within the 2\,s spin budget. Course~3 sits between these extremes: acute corners cause complete line loss similar to Course~1, but the compact triangle geometry keeps stored breadcrumbs within short travel distances, producing Stage-2 recoveries that complete quickly despite the heavier navigation overhead.

The TTR distributions (Figure~\ref{fig:ttr_hists}) reinforce this interpretation. Course~2 shows a bimodal distribution with a sharp sub-1\,s peak (Stage-1) and a 2--7\,s tail (Stage-2), the two populations separated by the 2\,s spin budget. Courses~1 and~3 show right-skewed unimodal distributions that peak in the Stage-2 range, confirming that Stage-1 contributes minimally when geometry forces complete occlusion. TTR therefore depends mainly on course geometry rather than on the specific fault, a finding that directly supports deployment planning for real-world line-following environments.

The 95\% Wilson confidence interval for the overall success rate (0.79--0.92) reflects the relatively modest episode count of 119. Widening the evaluation to a larger number of courses and fault conditions would tighten this interval. Nevertheless, the consistent within-course behavior and the clear geometry-driven mechanism split already provide strong evidence that the two-stage strategy is valid and that its performance characteristics are predictable rather than incidental.

\section{Threats to Validity}
\label{sec:threats}

\head{Simulation-to-reality gap.}
All experiments were conducted in Webots, which relies on idealized sensor models. In real deployments, cameras may be affected by motion blur, rolling shutter, glare, and calibration drift. Real depth sensors can also introduce flying pixels, quantization noise, and interference on reflective or IR-absorbing surfaces. These effects may distort the floor-plane estimate, create false obstacle detections, or hide genuine ones, which could reduce tracking and recovery performance in
physical deployment.

\head{Environment assumptions.}
The method assumes a clearly visible painted line on a mostly flat surface. Worn markings, shadows, textured tiles, or uneven lighting may weaken the HSV line mask. The per-row EMA floor model and SVD plane fitting also work best on smooth surfaces; ramps, thresholds, or rough terrain can violate the
planar assumption and lead to incorrect floor-versus-obstacle classification.

\head{VO accuracy.}
The lightweight monocular VO does not include loop closure, IMU fusion, or learned feature descriptors. During pure rotations and in low-texture corridors, drift can build up and shift Stage-2 navigation targets by 10--20\,cm, which may be enough to miss the line on a narrow track.

\head{Sample size.}
Although the 119-episode dataset yields confidence intervals that remain above the 0.79 lower bound, it is still relatively small compared with large-scale deployed fleet data. Results may differ across substantially different floor materials, line widths, or lighting conditions.

\section{Future Work}
\label{sec:future}

\head{Improved localization:} Replacing a lightweight monocular VO with a keyframe-based visual SLAM system (for example, ORB-SLAM3~\cite{campos2021:orbslam3}) would introduce loop closure and long-range drift correction, directly addressing failure category~(iii). Visual-inertial odometry (for example, VINS Mono~\cite{qin2018:vinsmono}) would further stabilize estimates during the pure-rotation phases of Stage-1 spin.

\head{Appearance-based global relocalization:} Adding a compact place-recognition front end (such as Galvez's bags or a learned global descriptor) would let Stage-2 propose candidate poses globally instead of routing to the nearest stored breadcrumb. This would address failure category~(i) by removing the dependence on local breadcrumb coverage.

\head{Adaptive recovery policy:} Replacing fixed-duration spin and navigation timeouts with a decision policy that balances expected TTR, perception confidence, and remaining breadcrumb quality would make recovery more efficient and responsive to runtime conditions.

\head{Integration with motion planners:} The current pivot-based obstacle avoidance could be replaced by exporting free-space estimates to a local cost map and connecting them to established planners (DWA, TEB) for smoother navigation, especially during Stage-2 recovery in cluttered environments.

\head{Formal self-adaptive verification:} The two-stage structure, with explicit budgets, timeouts, and confirmation thresholds, is well suited to formal analysis with probabilistic model checkers such as PRISM~\cite{kwiatkowska2011:prism,calinescu2011:dynamic}. Modeling the recovery protocol as a discrete Markov decision process would make it possible to derive formal bounds on self-healing success probability and expected TTR, then verify them against the empirical results reported here. This would strengthen
confidence in deployment scenarios not covered by simulation.

\head{Learning-based adaptive recovery policy:}  The fixed spin and navigation timeouts could be replaced with a reinforcement learning policy trained on MAPE-K feedback signals (line confidence, breadcrumb quality, remaining budget, and course geometry). Such a policy would adapt the self-healing behavior to environment-specific failure patterns observed at runtime, moving beyond the hand-tuned thresholds of the current design toward a genuinely self-adaptive controller~\cite{delemos2013:software}.

\head{Physical validation:} Deploying and validating the algorithm on a real JetBot or equivalent platform with a physical Orbbec Astra camera is the most important next step. Results on real hardware, under varied lighting, flooring materials, and line-wear conditions, would quantify the sim-to-real gap and confirm whether the self-healing behavior observed in simulation transfers to physical deployment.

\section{Conclusions}
\label{sec:conclusion}

This paper presents a lightweight, camera-first self-healing controller for autonomous ground vehicles that recovers from a complete loss of guidelines using only an RGB-D camera and CPU-based computation. The proposed system embeds a MAPE-K loop within a single 50\,ms control cycle, enabling perception, decision-making, and actuation to operate continuously without operator intervention or external adaptation infrastructure. Its design combines three tightly coupled components: a depth-gated HSV line tracker with online hue adaptation and periodic floor modeling, a depth-fused obstacle avoidance module that integrates geometric residuals with YOLOv8n detections, and a two-stage recovery mechanism that first performs in-place spin-and-search and then resorts to VO-guided navigation toward stored breadcrumb poses when immediate reacquisition fails.

The experimental evaluation on 119 fault-injected episodes across three geometrically diverse Webots simulation courses demonstrated that the system recovered successfully in 86.6\% of cases, with a 95\% Wilson confidence interval of 0.79--0.92. 
The median time to recovery was 3.26\,s, and the mean was 3.48\,s, indicating that robust recovery can be achieved under strict sensing and computational constraints. 
The results further show that recovery behavior is strongly shaped by environment geometry: partial-visibility failures are typically resolved rapidly by Stage~1 spin recovery, whereas complete-occlusion cases require Stage~2 navigation and incur longer recovery times. 
The post-recovery stabilization step also proved important in reducing oscillatory re-entry into recovery.

Overall, the findings show that reliable and repeatable visual recovery is feasible without LiDAR, GPS, or GPU-class hardware. Compared with prior heavier sensor-based approaches, the proposed method achieves competitive recovery performance while remaining low-cost, modular, and deployable on resource-constrained platforms. These results show that self-healing recovery is achievable on camera-only UGVs, and they motivate future work on real-world validation, improved localization, and more adaptive recovery policies.

\section*{Acknowledgment}

\noindent 
The first author carried out this work as part of a Master's project with the Secure and Trustworthy Intelligent Systems group at Simula Research Laboratory.

\appendix
\section{Parameter Settings}
\label{app:params}

Table~\ref{tab:params} shows the values of key system parameters used across all experiments.

\begin{table}[t]
\centering\floatfont
\caption{Key system parameters used across all experiments.}
\label{tab:params}\setlength{\tabcolsep}{3pt}
\vspace*{-1ex}\resizebox{\linewidth}{!}{%
\begin{tabular}{@{}lrl@{}}
\toprule
\thead[l]{Parameter} & \thead[r]{Value} & \thead[l]{Description} \\
\midrule
\multicolumn{3}{@{}l}{\emph{Depth-gated line tracker (Eq.~\ref{eq:ema}--\ref{eq:mask})}}\\
$\delta_h$ (recovery)          & $+8$     & Extra hue expansion in recovery \\
$\alpha_g$ (floor EMA)         & 0.05     & Per-row floor EMA weight \\
Ground warmup                  & 20       & Init frames for floor model \\
PLANE\_FIT\_EVERY\_N           & 3        & SVD plane-fit cadence (frames) \\
Floor residual / gradient      & 0.02\,m / 0.04 & Floor-pixel acceptance thresholds \\
Far-mask area / speed          & 500\,px / 0.65 & Far-mask trigger and speed scale \\
\midrule
\multicolumn{3}{@{}l}{\emph{Control loop and steering (Eq.~\ref{eq:pid}, \ref{eq:speed_scaling})}}\\
TIME\_STEP / $\Delta t$        & 50\,ms / 0.05\,s & Control loop period (20\,Hz) \\
$K_p$                          & 0.9      & PD steering proportional gain \\
$K_d$                          & 0.1      & PD steering derivative gain \\
\midrule
\multicolumn{3}{@{}l}{\emph{Depth-fused obstacle detection and avoidance}}\\
$k_{\mathrm{mad}}$             & 2.0      & MAD residual scale \\
$\tau_{\mathrm{margin}}$       & 0.05\,m  & Extra margin over floor model \\
Min width / area / persist     & 0.03\,m / 140\,px / 2 & Obstacle size and persistence filters \\
$d_{\mathrm{warn}}$            & 1.40\,m  & Obstacle warning distance \\
$d_{\mathrm{near}}$            & 0.75\,m  & Obstacle near / pivot distance \\
\midrule
\multicolumn{3}{@{}l}{\emph{Breadcrumb navigation (Eq.~\ref{eq:nav})}}\\
$T_{\mathrm{step}}$            & 8.0\,s   & Max navigate-to-breadcrumb time \\
\midrule
\multicolumn{3}{@{}l}{\emph{Two-stage recovery}}\\
$N_{\mathrm{confirm}}$         & 5        & Consecutive confirmation frames \\
$T_{\mathrm{spin}}$            & 2.0\,s   & Max spin-search duration \\
$N_{\mathrm{max}}$             & 3        & Max spin--navigate cycles \\
Grace window                   & 0.6\,s   & Delay before declaring loss again \\
Stabilization hold             & 0.6\,s   & Hold still after reacquisition \\
Post-recovery relax            & 0.8\,s   & Looser thresholds after recovery \\
\midrule
\multicolumn{3}{@{}l}{\emph{Detector}}\\
YOLO input                     & 640$\times$640 & YOLOv8n input resolution \\
YOLO\_CONF\_THRESH             & 0.35     & Confidence threshold \\
YOLO\_NMS\_THRESH              & 0.45     & NMS IoU threshold \\
\bottomrule
\end{tabular}}
\end{table}

\printbibliography
 \EOD

\end{document}